\documentclass[preprint,12pt]{elsarticle}

\usepackage{amssymb}
\usepackage{amsmath}
\usepackage{amsthm}
\usepackage{hyperref}
\usepackage{multirow}
\usepackage{subcaption}
\usepackage{booktabs}

\newtheorem{theorem}{Theorem}
\newtheorem{lemma}{Lemma}

\journal{Information Fusion}

\begin{document}

\begin{frontmatter}



\title{Directed Homophily-Aware Graph Neural Network}




\author[label1]{Aihu Zhang}
\author[label1]{Jiaxing Xu}
\author[label1]{Mengcheng Lan}
\author[label2]{Shili Xiang}
\author[label1]{Yiping Ke}

\affiliation[label1]{organization={College of Computing and Data Science, Nanyang Technological University},
         addressline={50 Nanyang Avenue},
         postcode={639798},
         country={Singapore}}

\affiliation[label2]{organization={Institute for Infocomm Research, A*STAR},
             addressline={2 Fusionopolis Way, Innovis 08-01},
             postcode={138634},
             country={Singapore}}

\begin{abstract}
Graph Neural Networks (GNNs) have achieved significant success in various learning tasks on graph-structured data. Nevertheless, most GNNs struggle to generalize to heterophilic neighborhoods. Additionally, many GNNs ignore the directional nature of real-world graphs, resulting in suboptimal performance on directed graphs with asymmetric structures. In this work, we propose Directed Homophily-aware Graph Neural Network (DHGNN), a novel framework that addresses these limitations by incorporating homophily-aware and direction-sensitive components. DHGNN employs a resettable gating mechanism to adaptively modulate message contributions based on homophily levels and informativeness, and a structure-aware noise-tolerant fusion module to effectively integrate node representations from the original and reverse directions. Extensive experiments on both homophilic and heterophilic directed graph datasets demonstrate that DHGNN outperforms state-of-the-art methods in node classification and link prediction. In particular, DHGNN improves over the best baseline by up to 15.07\% in link prediction. Our analysis further shows that the gating mechanism captures directional homophily gaps and fluctuating homophily across layers, providing deeper insights into message-passing behavior on complex graph structures.
\end{abstract}






\end{frontmatter}



\begingroup
\renewcommand\thefootnote{}
\footnotetext{Our code is available at: \url{https://anonymous.4open.science/r/DHGNN-7FE5}.}
\addtocounter{footnote}{-1}
\endgroup

\section{Introduction}

Graph Neural Networks (GNNs)~\cite{gnn2009, gcn2016, gat2017, sage2017} are powerful models for learning graph representations, typically following a message passing paradigm~\cite{gilmer2017mp}, where nodes iteratively aggregate and combine messages from their neighbors. After several iterations, each node obtains an embedding that captures information from itself and nearby nodes, supporting various node- and graph-level tasks~\cite{khemani2024review}. Despite their success, message-passing GNNs often struggle to generalize in two settings: (1) heterophilic neighborhoods, and (2) directed graphs.

\begin{figure*}[htbp]
    \centering
    \begin{subfigure}[t]{0.45\textwidth}
        \centering
        \includegraphics[width=\linewidth]{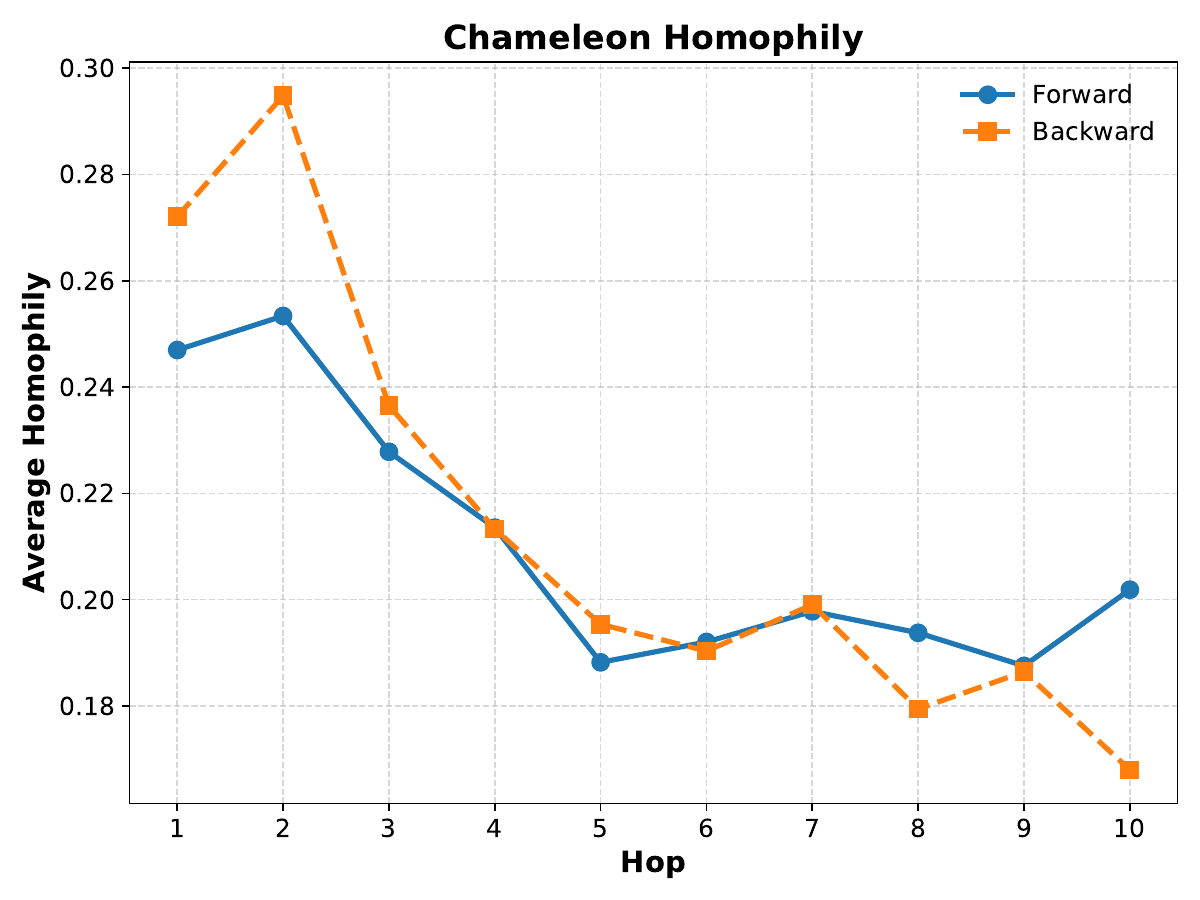}
        \caption{The homophily fluctuation in Chameleon.}
        \label{subfig:chameleon}
    \end{subfigure}
    \hfill
    \begin{subfigure}[t]{0.45\textwidth}
        \centering
        \includegraphics[width=\linewidth]{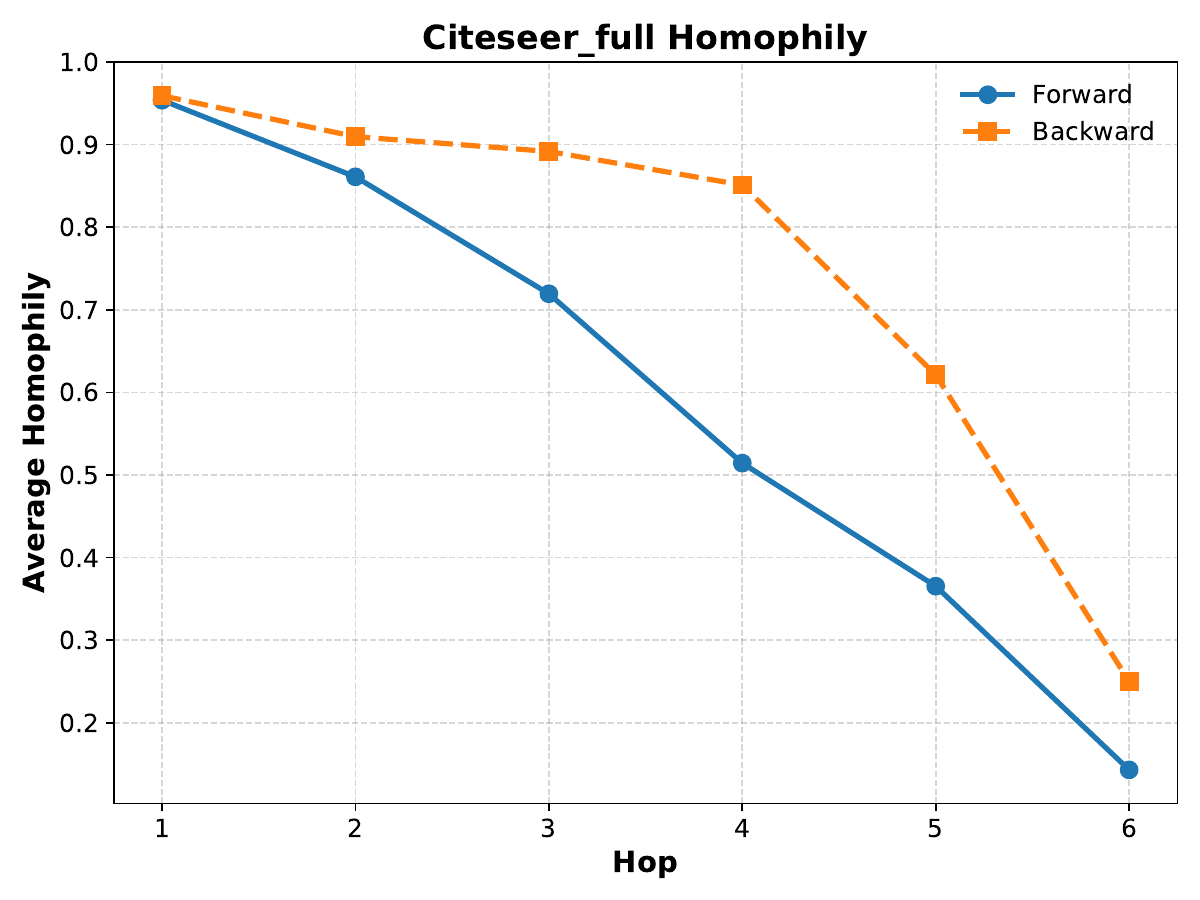}
        \caption{The directional homophily difference in Citeseer-Full.}
        \label{subfig:citeseer}
    \end{subfigure}
    \caption{The average node-wise homophily ratio in different hops. Forward denotes the direction in which the graph is constructed while backward denotes the reverse direction. The line chart reveals (a) the homophily ratio does not consistently decrease with increasing distance, and (b) the homophily levels of forward and backward neighbors can be different. We further explore this phenomenon with an example in \ref{app:homoanalysis}.}
    \label{fig:homophily}
\end{figure*}

Heterophily refers to the case where neighboring nodes have dissimilar features or labels~\cite{zhu2020beyond,glognn2022, tang2023generalized, xie2025robust, guo2025learning}. The node-wise homophily ratio~\cite{pei2020geom} quantifies the proportion of a node’s neighbors sharing the same label, which is commonly used to measure heterophily. GNNs often perform poorly on graphs with low average homophily~\cite{h2gcn2020}, partly due to over-smoothing~\cite{li2018smooth}, where deeper models produce indistinguishable node representations. Several methods have been proposed to address heterophilic settings and mitigate over-smoothing~\cite{h2gcn2020, yan2022ggcn, suresh2021wrgnn, he2022bmgcn, luan2022acm, song2023ordered}. However, these approaches either fail to capture fluctuations in homophily across different hops~\cite{h2gcn2020, song2023ordered} (illustrated in Figure~\ref{subfig:chameleon}) or lack the ability to make personalized, node-specific decisions~\cite{suresh2021wrgnn, yan2022ggcn, he2022bmgcn, luan2022acm}, limiting their effectiveness in complex neighborhood structures.

\begin{figure*}[htbp]
    \centering
    \begin{subfigure}[t]{0.49\textwidth}
        \includegraphics[width=\linewidth]{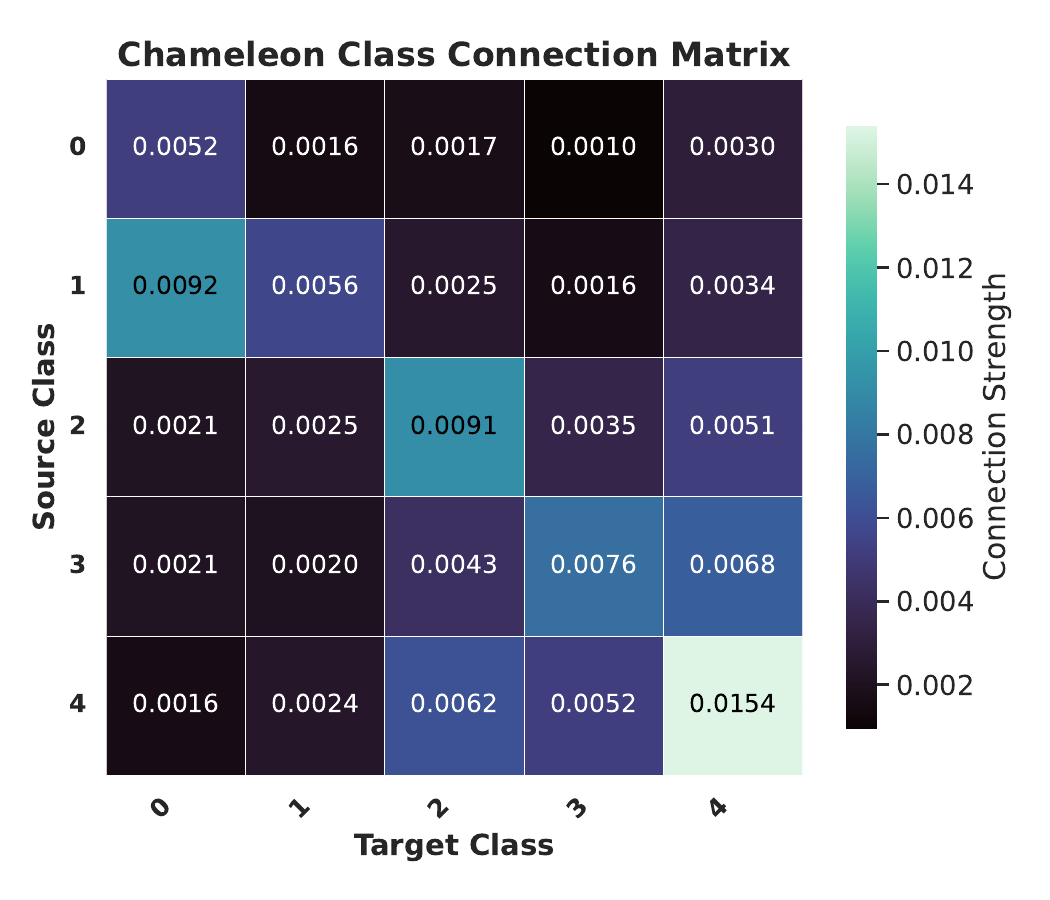}
        \caption{Directed Chameleon dataset.}
        \label{fig:chameleondir}
    \end{subfigure}
    \hfill
    \begin{subfigure}[t]{0.49\textwidth}
        \includegraphics[width=\linewidth]{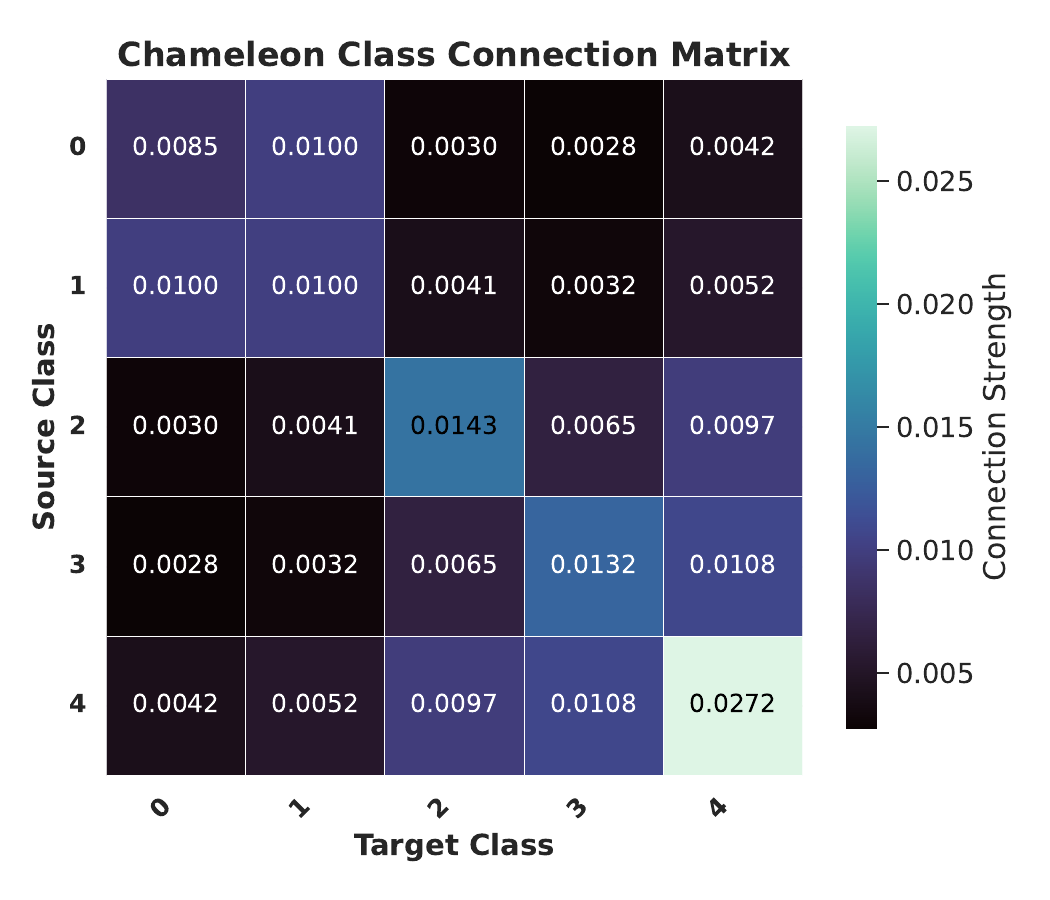}
        \caption{Undirected Chameleon dataset.}
        \label{fig:chameleonu}
    \end{subfigure}
    \caption{Class connection matrices of the directed and undirected versions of the same graph. The matrices tend to be asymmetric for heterophilic graphs, such as Chameleon. Converting this graph from directed to undirected could eliminate the asymmetry between the edge ratios from class 1 to class 0 and from class 0 to class 1, while the two ratios differ significantly in the original directed graph.}
    \label{fig:connection}
\end{figure*}

Edge directionality is another important source of information that GNNs fail to fully exploit. While the traditional approach of converting directed graphs to undirected ones~\cite{gcn2016, gat2017, sage2017} preserves intra-class relationships, it imposes artificial symmetry that can obscure important directional, inter-class information, as illustrated in Figure~\ref{fig:connection}. Both forward (original) and backward (reversed) directions can contain complementary and informative signals that are critical to understand complex graph structures, such as inter-domain connections in citation networks. Encoding the two directions separately demonstrates effectiveness~\cite{tong2020dgcn,dirgnn2024,nddgnn2024}. Nonetheless, these approaches still face limitations. They ignore the homophily and structure differences between neighborhoods in opposite directions, as shown in Figure~\ref{subfig:citeseer}. Moreover, their reliance on layer-wise fusion restricts the model’s ability to capture long-range dependencies across the graph~\cite{dirgnn2024,nddgnn2024}.

In this paper, we propose Directed Homophily-aware Graph Neural Network (DHGNN), a novel framework designed to extract meaningful information from complex, noisy neighborhoods by explicitly modeling homophily fluctuations across hops, and structural as well as homophily gaps between opposite directions.

To address the challenges posed by heterophilic neighborhoods, we introduce a homophily-aware gating mechanism that dynamically adjusts gating values for message-passing based on the embedding similarity and informativeness of neighboring nodes across different hops. This design enables the model to handle fluctuating homophily patterns and selectively incorporate relevant signals, even from distant neighbors.

To capture the structural and homophily differences between opposite directions, we design a dual-encoder architecture that separately processes the forward and backward directions of directed graphs. A structure-aware, noise-tolerant fusion mechanism then adaptively integrates the directional embeddings, guided by both node-level features and structural context, to form a unified and expressive node representation. By employing independent encoders along with branch-specific auxiliary losses, the model effectively captures distinct and meaningful information from each direction.

To validate the effectiveness of DHGNN, we conduct comprehensive experiments on five benchmark datasets with varying levels of homophily and directionality. DHGNN is able to outperform state-of-the-art baselines in node classification and link prediction tasks. Further analysis demonstrates its ability to adaptively respond to both the homophily fluctuations across hops and the directional homophily gaps. An analysis of DHGNN's behavior at deeper layers suggests that the model effectively alleviates the over-smoothing issue.

Our key contributions can be summarized as follows:

\begin{itemize}
    \item We propose a homophily-aware gating mechanism that adaptively regulates message-passing weights across layers and directions.
    \item We develop a structure-aware, noise-tolerant fusion strategy that integrates directional embeddings into a unified representation.
    \item We validate the proposed DHGNN on five heterophilic and homophilic datasets. The results demonstrate improved classification and link prediction performance by up to 15.07\%, along with interpretable gate behavior.
\end{itemize}

\section{Related Work}

\textbf{Graph Neural Networks for heterophilic graphs.} Various methods have been developed to handle heterophily in graphs. LINKX~\citep{linkx2021} avoids aggregating over heterophilic neighborhoods by separately processing node features and adjacency information. Other approaches decompose ego and neighbor representations or utilize signal separation techniques~\citep{h2gcn2020, suresh2021wrgnn, he2022bmgcn, glognn2022, luan2022acm}. For example, H2GCN~\citep{h2gcn2020} concatenates information from different neighborhood ranges while excluding self-loops. WRGNN~\citep{suresh2021wrgnn} and BM-GCN~\citep{he2022bmgcn} apply different transformation matrices to ego and neighbor messages. GloGNN~\citep{glognn2022} uses both low- and high-pass filters, while ACM-GCN~\citep{luan2022acm} distinguishes ego embeddings using identity filters at the channel level.

Several methods employ adaptive weighting strategies. GPR-GNN~\citep{gprgnn2020} learns Generalized PageRank weights across layers. GGCN~\citep{yan2022ggcn} uses learnable scalars for self and signed neighbor edges. Gradient Gating ($G^2$)\citep{gradientgating2022} introduces multi-rate gradient flow, and Ordered GNN~\citep{song2023ordered} applies soft gating to gradually freeze node embeddings and mitigate over-smoothing. However, most gating-based approaches assume a monotonically decreasing homophily with distance, overlooking potential fluctuations, which may limit the ability to extract informative signals from distant nodes.

\textbf{Directed Graph Neural Networks.} Traditional GNNs often convert directed graphs into undirected ones, potentially disrupting important inter-class links. To address this limitation, various methods have been developed from both spectral and spatial perspectives to explicitly model directionality in graph learning. Spectral approaches include DiGCN~\citep{digcn2020}, which approximates the Laplacian using personalized PageRank transition probabilities, and MagNet~\citep{magnet2021}, which introduces a Magnetic Laplacian with complex-valued phase matrices to encode directionality. Building on this, \citep{geisler2023transformers} uses both the Magnetic Laplacian and directional random walk encodings as positional encodings for Transformers, while \citep{huang2024posenc} further enhances the Magnetic Laplacian-based encoding with additional structural factors. However, these methods often apply the same transformation to both edge directions, overlooking directional differences in neighborhood homophily.


Spatial approaches explicitly separate directional representations. DGCN~\citep{tong2020dgcn} uses distinct functions for first- and second-order proximity. Dir-GNN~\citep{dirgnn2024} independently encodes each direction and combines them with manually set fusion weights, while NDDGNN~\citep{nddgnn2024} adaptively learns fusion weights based on structural and feature similarities. Although effective for local interactions, such layer-wise fusion may struggle to capture long-range dependencies along extended paths.

\section{Preliminaries}

\textbf{Directed graphs.} In this paper, we address the tasks of semi-supervised node classification and link prediction on attributed directed graphs. Given a directed graph $\mathcal{G} = (\mathcal{V}, \mathbf{X}, \mathbf{A})$ with $n$ nodes and $m$ edges, each node $i \in \mathcal{V}$ is associated with a feature vector, forming a feature matrix $\mathbf{X} \in \mathbb{R}^{n \times d}$, where $d$ is the feature dimension. The graph also includes a label $y_i \in \{1, \dots, C\}$ for each node corresponding to $C$ classes. The directed structure is represented by an adjacency matrix $\mathbf{A} \in \{0, 1\}^{n \times n}$, where $\mathbf{a}_{ij} = 1$ if there is a directed edge from node $i$ to node $j$, and zero otherwise. 

\textbf{Homophily.} Homophily is the tendency of connected nodes in a graph to share similar labels or features, and it underpins the success of many GNNs. A common way to measure this is the node-wise homophily ratio, defined for a node $v$ with label $y_v$ as
\begin{equation}
\label{eq:homo}
h_v = \frac{\left| \left\{ u \in \mathcal{N}(v) \;\middle|\; y_u = y_v \right\} \right|}{\left| \mathcal{N}(v) \right|},
\end{equation}
where $\mathcal{N}(v)$ is the set of neighbors of $v$. The overall graph homophily is computed by averaging $h_v$ over all nodes. High homophily typically benefits standard GNNs, while low homophily (high heterophily) poses challenges for conventional message passing.

In scenarios where node labels are partially or entirely unavailable, such as in link prediction tasks, the notion of homophily can be extended to quantify feature similarity between neighboring nodes.

\textbf{Message-passing neural networks.} Most GNNs follow a message passing mechanism~\citep{xu2018powerful} iteratively updating node representations of node $v$, namely $\mathbf{{h}}^{(l)}_v \in \mathbb{R}^{p}$, where $l$ denotes the layer number and $p$ is the hidden dimension at current layer. Each layer of a message passing GNN comprises two fundamental steps, i.e.,  aggregation and combination, which can be formulated as:
\begin{equation}
\label{eq:mpnn}
    \begin{aligned}
        \mathbf{m}^{(l)}_v &= \text{AGGREGATE}^{(l)} (\{\mathbf{h}^{(l-1)}_u | u \in  \mathcal{N}(v)\}), \\
        \mathbf{h}^{(l)}_v &= \text{COMBINE}^{(l)} (\mathbf{h}^{(l-1)}_v, \mathbf{m}^{(l)}_v),
    \end{aligned}
\end{equation}
where the $\text{AGGREGATE}(\cdot)$ takes the embeddings of nodes in the neighborhood $\mathcal{N}(v)$ of a node $v$ and generates a message $\mathbf{m}^{(l)}_v$, and $\text{COMBINE}(\cdot)$ incorporates the message with the previous node embedding $\mathbf{h}^{(l-1)}_v$ to produce the updated embedding $ \mathbf{h}^{(l)}_v$. The initial embedding $\mathbf{h}^{(0)}_v$ is set to be the original node feature $\mathbf{x}_v$ of node $v$.

\section{Methodology}
\label{sec:method}

\subsection{Model architecture}

\begin{figure*}[htbp]
    \centering
    \includegraphics[width=\linewidth]{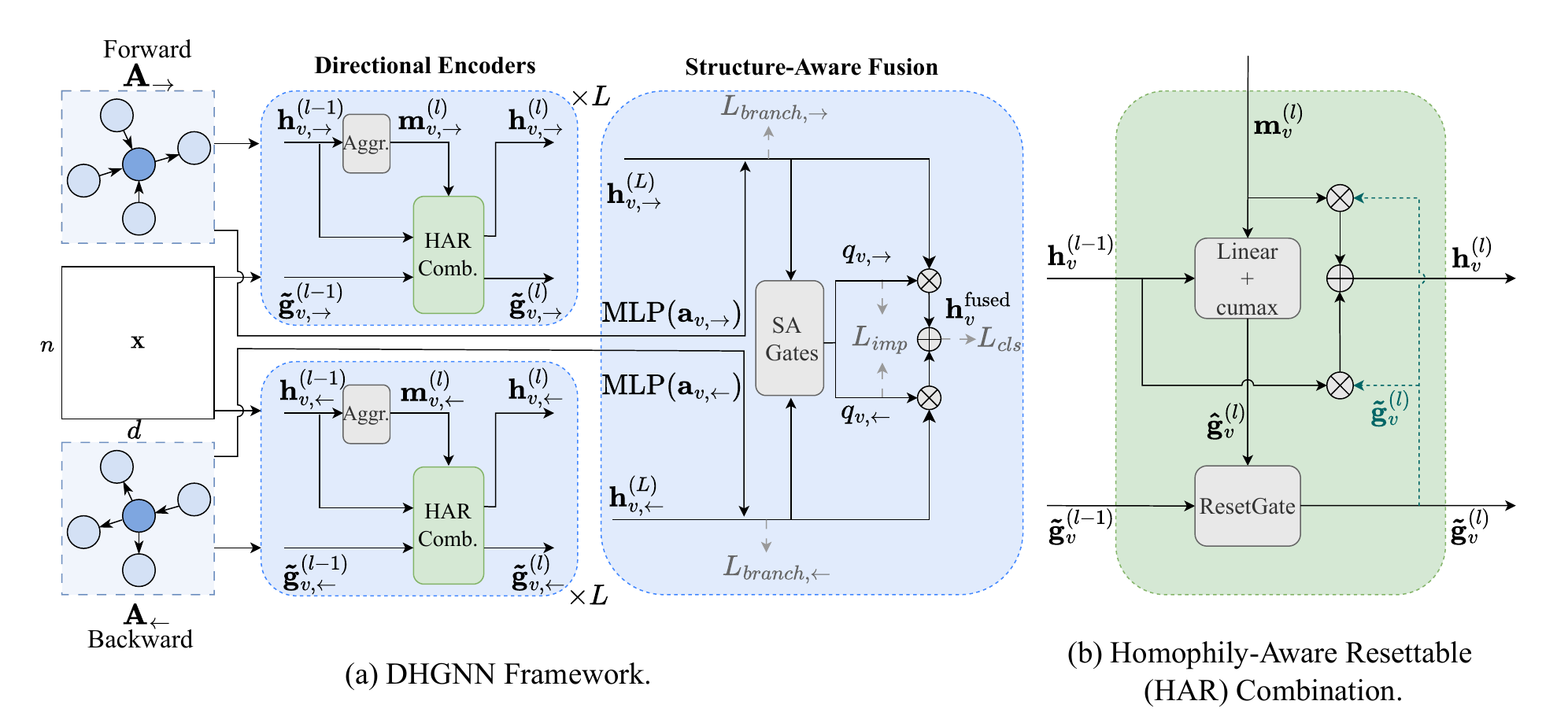}
    \caption{The framework of DHGNN. ``SA Gates'' is short for the structure-aware noise-tolerant gates. The model consists of two message-passing modules (one for each edge direction) and a fusion module. Node representations are learned independently for the forward and backward directions, and subsequently fused into a unified representation.}
    \label{fig:framework}
\end{figure*}

To enable the model to effectively handle varying levels of homophily in the forward and backward directions, we propose a novel and effective framework specifically designed to disentangle and adaptively integrate directional information, namely DHGNN as shown in Figure~\ref{fig:framework}a. 

The model employs two parallel encoders based on a resettable gated message-passing scheme. The two encoders share the same architecture but with independent parameters, which are responsible for processing forward and backward edges, respectively. Within each encoder, a gating vector dynamically regulates the contribution of incoming messages. The resettable gating mechanism is designed to attenuate or amplify messages based on their informativeness, particularly preserving informative long-range dependencies. This enables the model to propagate relevant information more effectively across multiple hops.


The directional embeddings are then combined using a structure-aware fusion mechanism. Fusion gating values are derived from both directional embeddings and structural cues from the directed adjacency matrices, allowing the model to integrate directional information based on graph topology.

To avoid the collapse of the two directional encoders into a single dominant pathway, DHGNN introduces an auxiliary importance loss that discourages extreme imbalance in the fusion decision scores. Furthermore, to encourage disentanglement and minimize redundancy between the forward and backward encoders, a branch classification loss is incorporated during training.


\subsection{Homophily-aware gated message passing}

As presented in the previous sections, the homophily ratio could vary in the backward and forward directions, and the homophily ratio is not strictly decreasing in all datasets. 

Therefore, to capture fluctuations in local homophily and gaps between forward and backward directions simultaneously, we introduce a homophily-aware gated message-passing mechanism, as defined in Eq. (\ref{eq:mp}). In each aggregation layer, a separate gating vector is updated with a homophily-aware resettable
(HAR) combination module alongside the node embeddings (shown in Figure~\ref{fig:framework}b), modulating the influence of forward messages from the neighborhood in a node-specific manner.
\begin{equation}
\label{eq:mp}
\begin{split}
\mathbf{m}_v^{(l)} &= \text{mean} \left( \left\{ \mathbf{h}_u^{(l-1)} | u \in \mathcal{N}(v) \right\} \right), \\
\mathbf{\hat{g}}_v^{(l)} &= \text{cumax}_{\rightarrow} \left( \mathbf{W}^{(l)} \left[ \mathbf{h}_v^{(l-1)} || \mathbf{m}_v^{(l)} \right] + \mathbf{b}^{(l)} \right), \\
\mathbf{\tilde{g}}_v^{(l)} &= \text{ResetGate} (\mathbf{\tilde{g}}_v^{(l-1)}, \mathbf{\hat{g}}_v^{(l)}), \\
\mathbf{h}_v^{(l)} &= \mathbf{\tilde{g}}_v^{(l)} \circ \mathbf{h}_v^{(l-1)} + (1 - \mathbf{\tilde{g}}_v^{(l)}) \circ \mathbf{m}_v^{(l)},
\end{split}
\end{equation}
where $\text{cumax}_{\rightarrow}(\cdot) = \text{cumsum}_{\rightarrow}(\text{softmax}(\cdot))$ denotes a composite operation that applies the softmax function followed by a cumulative sum. The subscript $\rightarrow$ indicates that the accumulation proceeds in a left-to-right direction. $\mathbf{W}^{(l)}$ and $\mathbf{b}^{(l)}$ are the learnable weight and bias term in the multilayer perceptron (MLP) layer, $||$ stands for concatenation, and $\circ$ denotes element-wise multiplication. The first equation represents the aggregation step in the message-passing layer, while the remaining three equations define the combination step.

The gating vector is updated using a Gated Recurrent Unit (GRU)-inspired resettable gating mechanism, which enables selective resetting of its values. The reset gate $\mathbf{r}_v^{(l)}$ is computed via an MLP conditioned on the current preliminary gating vector $\mathbf{\hat{g}}_v^{(l)}$ and the previous gating vector $\mathbf{\tilde{g}}_v^{(l-1)}$:
\begin{equation}
\label{eq:rg}
\begin{split}
\mathbf{\tilde{g}}_v^{(l)} &= \text{ResetGate} (\mathbf{\tilde{g}}_v^{(l-1)}, \mathbf{\hat{g}}_v^{(l)}) = \left(1 - \mathbf{r}_v^{(l)}\right) \circ \mathbf{\tilde{g}}_v^{(l-1)} + \mathbf{r}_v^{(l)} \circ \mathbf{\hat{g}}_v^{(l)}, \\
\mathbf{r}_v^{(l)} &= \sigma (\mathbf{W}_r \mathbf{\hat{g}}_v^{(l)} + \mathbf{U}_r \mathbf{\tilde{g}}_v^{(l-1)} + \mathbf{b}_r),
\end{split}
\end{equation}
where $\sigma(\cdot)$ stands for the sigmoid function, and $\mathbf{W}_r$, $\mathbf{U}_r$ and $\mathbf{b}_r$ are the learnable weights and the bias term, respectively. The values in the gating vector tend to increase when the forward neighborhood messages deviate from the current node embedding. These values are constrained within the range $[0,1]$, where higher values correspond to a reduced influence from neighborhood messages at the current layer. This selective suppression of information from heterophilic neighbors helps mitigate the over-smoothing issue commonly observed in heterophilic graphs.

While the gating strategy at the combination step is inspired by Ordered GNN~\citep{song2023ordered}, the original design assumes that the neighborhood homophily ratio strictly decreases with increasing hop distance. In contrast, our gating mechanism is designed to accommodate fluctuations in neighborhood homophily across different hops. To this end, the gating vector values are not constrained to follow a strictly monotonic trend across layers. This behavior is enabled by the resettable gating mechanism, which allows gating values to increase when the homophily ratio we infer from the embedding similarity decreases or when the incoming message is uninformative, and decrease otherwise. Such flexibility enables the model to gradually attenuate less relevant signals from distant nodes, while still retaining useful long-range information when necessary. As a result, DHGNN can dynamically adapt its receptive field depending on the directionality and structure of the graph, leading to more expressive and context-aware node representations. We provide an analysis of the monotonicity of gating values in \ref{app:monoanalysis}.

To prevent premature saturation of gating vectors during message passing and reduce the parameter count in the gating network, we employ a chunking strategy similar to the one in~\citep{song2023ordered}. 

\subsection{Structure-aware noise-tolerant fusion}

The representations learned from the forward and backward directions can differ significantly due to variations in neighborhood homophily and linking patterns. To fully exploit the information provided by each direction, we design a structure-aware noise-tolerant fusion mechanism. This mechanism integrates the directional node embeddings generated by two separate encoders, producing a unified representation before it is passed to downstream tasks. The fusion process is formulated as follows:
\begin{equation}
\label{eq:fus}
\begin{split}
\mathbf{h}_v^{\text{fused}} = \sigma \left(q_{v,\leftarrow} \mathbf{h}^{(L)}_{v,\leftarrow} + q_{v,\rightarrow} \mathbf{h}^{(L)}_{v,\rightarrow}\right),
\end{split}
\end{equation}
where $\sigma(\cdot)$ is a linear layer, scalars $q_{v,\leftarrow}$ and $q_{v,\rightarrow}$ indicate the decision scores for each direction, and $L$ denotes the number of encoder layers. These scores are given by a noise-tolerant MLP-based gating mechanism (To simplify the notation, we omit the layer index $L$ and use only a single-direction subscript in the following equations, as the processing steps for the two directions are analogous.):
\begin{equation}
\label{eq:fusdetail}
\begin{split}
q_{v,\leftarrow}, q_{v,\rightarrow} &= \text{Softmax}\left(\text{NTB}(\mathbf{\tilde{h}}_{v,\leftarrow}), \text{NTB}(\mathbf{\tilde{h}}_{v,\rightarrow})\right),\\
\mathbf{\tilde{h}}_{v,\leftarrow} &= \mathbf{h}_{v,\leftarrow} + \beta \text{MLP} (\mathbf{a}_{v,\leftarrow}) ,\\ 
\text{NTB}(\mathbf{\tilde{h}}_{v,\leftarrow}) &= \mathbf{\tilde{h}}_{v,\leftarrow} \mathbf{W}_{1,\leftarrow} + \epsilon \cdot \text{Softplus}(\mathbf{\tilde{h}}_{v,\leftarrow} \mathbf{W}_{2,\leftarrow}),
\end{split}
\end{equation}
where $\text{NTB}(\cdot)$ stands for the noise-tolerant block, $\mathbf{a}_{v}$ is the corresponding vector in $\mathbf{A}$ containing all outgoing neighbors of $v$, and $\epsilon \sim N(0, 1)$ denotes standard Gaussian noise. We introduce this noise term in the fusion mechanism to promote a stable and decisive combination of the two directional representations. $\mathbf{W}_{1,\leftarrow}, \mathbf{W}_{2,\leftarrow} \in \mathbb{R}^{p \times 1}$ are learnable weights that control clean and noisy scores, respectively. Since the adjacency matrix of a directed graph is asymmetric, it can provide richer structural information than its undirected counterpart. To fully leverage this asymmetric structure in guiding the fusion process, we incorporate an adjacency-based embedding, which has been proven effective as a structural embedding in LINKX~\citep{linkx2021}, into the input of $\text{NTB}(\cdot)$. This integration is governed by the hyperparameter $\beta$.

To prevent two encoders from collapsing into one, we incorporate an importance loss following \citep{wang2023graph}:

\begin{equation}
\label{eq:imloss1}
\begin{split}
\text{Importance}(\mathbf{h}_{\leftarrow}) &= \sum_{v \in \mathcal{V}}  q_{v,\leftarrow}, \\
\text{Importance}(\mathbf{h}_{\rightarrow}) &= \sum_{v \in \mathcal{V}}  q_{v,\rightarrow}, \\
L_{imp} = \text{CV}(\text{Importance}(\mathbf{h}_{\leftarrow})
&, \text{Importance}(\mathbf{h}_{\rightarrow}))^2,
\end{split}
\end{equation}
where the importance score, $\text{Importance}(\cdot)$, is defined as the sum of decision scores $q_{v,\leftarrow}$ and $q_{v,\rightarrow}$ across all nodes in the same direction. Here, $\text{CV}(\cdot)$ denotes the coefficient of variation. Consequently, the importance loss $L_{imp}$ quantifies the variability of these importance scores, encouraging all encoders to maintain a comparable level of importance.

\subsection{Objective function}

The final objective function consists of three components. The primary component is the classification loss $L_{cls}$, computed using the fused node embedding $\mathbf{h}_v^{\text{fused}}$. We adopt focal loss for node classification and cross-entropy loss for link prediction.

The first auxiliary component is the importance loss $L_{imp}$, as described in the previous section. The second auxiliary component, the branch classification loss $L_{branch}$, encourages each branch to independently learn information useful for node classification. It is calculated as the average of classification losses from each directional branch. Formally, $L_{branch}=(L_{branch,\leftarrow} + L_{branch,\rightarrow})/2$. We adopt cross-entropy loss for both directions.

To prevent the fusion mechanism from interfering with the individual optimization of each directional encoder, we apply stop-gradient operations to the directional embeddings after computing their respective losses and before they are fused into the final output. The stop-gradient operation effectively prevents gradients from propagating back through the encoder for each branch, ensuring that the previous quantities (such as the node embeddings from each encoder) are treated as fixed targets rather than learnable inputs during the update step. The two auxiliary losses are weighted by hyperparameters $\lambda_1$ and $\lambda_2$, respectively. $\lambda_2$ is introduced to form the total loss as a convex combination of the losses from the fused output and the individual branches. Thus, the overall objective function is formulated as:
\begin{equation}
\label{eq:objective}
\begin{split}
L_{total} = (1-\lambda_2)(L_{cls} + \lambda_1 L_{imp}) + \lambda_2 L_{branch}.
\end{split}
\end{equation}
\section{Experiments}
\label{sec:experiments}

\subsection{Datasets}


We evaluate our model on five representative benchmarks. Cora-ML and Citeseer-Full~\citep{bojchevski2017deep} are homophilic citation networks whose nodes represent scientific publications with category annotations and edges denote citation links. Chameleon and Squirrel~\citep{pei2020geom} are heterophilic Wikipedia page–page networks where nodes are articles connected by mutual links, equipped with bag-of-nouns features and traffic-based labels. Roman-Empire~\citep{platonov2023critical} is a word-level heterophilic graph constructed from a long Wikipedia article, where nodes are tokens linked by textual adjacency or dependency relations, with FastText-based features~\citep{grave2018learning} and syntactic-role labels. Additional dataset details are provided in Table~\ref{tab:data}.

\begin{table*}[htbp]
\centering
\footnotesize
\caption{Dataset properties and statistics.}
\begin{tabular}{l|rrrrr}
\toprule
 & \textbf{Cora-ML} & \textbf{Citeseer-Full} & \textbf{Chameleon} & \textbf{Squirrel} & \textbf{Roman-Empire} \\
\midrule
\#Nodes    & 2,995  & 4,230  & 2,277  & 5,201   & 22,662   \\
\#Edges    & 8,416  & 5,358  & 36,101  & 217,073   & 44,363   \\
\#Features & 2,879  & 602  & 2,325  & 2,089  & 300 \\
\#Classes  & 7      & 6      & 5      & 5     &  18 \\
Edge Hom.  & 0.792   & 0.949   & 0.235   & 0.223  & 0.500  \\
\bottomrule
\end{tabular}
\label{tab:data}
\end{table*}

\subsection{Experimental setup}

The baseline models used for comparison fall into three categories: (1) Traditional undirected graph methods, such as GCN~\citep{gcn2016}, and GAT~\citep{gat2017}; (2) State-of-the-art undirected graph methods for heterophilic graphs, including H2GCN~\citep{h2gcn2020}, GPRGNN~\citep{gprgnn2020}, LINKX~\citep{linkx2021}, FSGNN~\citep{fsgnn2021}, ACM-GCN~\citep{luan2022acm}, GloGNN~\citep{glognn2022}, and $G^2$ (Gradient Gating)~\citep{gradientgating2022}; and (3) State-of-the-art directed graph methods, comprising DiGCN~\citep{digcn2020}, MagNet~\citep{magnet2021}, Dir-GNN~\citep{dirgnn2024}, and NDDGNN~\citep{nddgnn2024}. 

For those baselines with results reported in \citep{nddgnn2024}, we adopt the results in the paper. For DHGNN, we select the best hyperparameters from 100 runs using the Optuna hyperparameter optimization framework~\citep{akiba2019optuna}. The selected hyperparameters are detailed in \ref{app:set}.

We conduct our experiments on an Intel Core i9-10940X CPU and 2 NVIDIA RTX A5000 GPUs.

\subsection{Node classification}

\begin{table}[htbp]
\centering
\footnotesize
\caption{Node classification accuracy (\%) across homophilic and heterophilic datasets. The best results are highlighted in \textbf{bold}, while the second-best results are marked with \underline{underline}.}
\setlength{\tabcolsep}{4pt}
\renewcommand{\arraystretch}{1.2}
\begin{tabular}{l|ccccc}
\toprule
\textbf{Datasets} & \textbf{Cora-ML} & \textbf{Citeseer-Full} & \textbf{Chameleon} & \textbf{Squirrel} & \textbf{Roman-Empire} \\
\midrule
MLP & 77.48{\tiny$\pm$1.23} & 80.01{\tiny$\pm$1.23} & 46.36{\tiny$\pm$2.57} & 34.14{\tiny$\pm$1.94} & 65.76{\tiny$\pm$0.42} \\
GCN & 52.37{\tiny$\pm$1.77} & 54.65{\tiny$\pm$1.22} & 64.82{\tiny$\pm$2.24} & 53.43{\tiny$\pm$2.01} & 73.69{\tiny$\pm$0.74} \\
GAT & 54.12{\tiny$\pm$1.56} & 55.15{\tiny$\pm$1.31} & 45.56{\tiny$\pm$3.16} & 39.14{\tiny$\pm$2.88} & 71.16{\tiny$\pm$0.63} \\
\midrule
H2GCN & 62.86{\tiny$\pm$1.45} & 68.34{\tiny$\pm$1.36} & 59.39{\tiny$\pm$0.98} & 37.90{\tiny$\pm$0.02} & 60.11{\tiny$\pm$0.52} \\
GPRGNN & 68.88{\tiny$\pm$1.66} & 70.12{\tiny$\pm$1.24} & 62.85{\tiny$\pm$2.90} & 54.35{\tiny$\pm$0.87} & 64.85{\tiny$\pm$0.27} \\
ACM-GCN & 69.96{\tiny$\pm$1.53} & 73.61{\tiny$\pm$1.32} & 74.76{\tiny$\pm$2.20} & 67.40{\tiny$\pm$2.21} & 69.66{\tiny$\pm$0.62} \\
GloGNN & 73.78{\tiny$\pm$1.69} & 76.13{\tiny$\pm$1.14} & 57.88{\tiny$\pm$1.76} & 71.21{\tiny$\pm$1.84} & 59.63{\tiny$\pm$0.69} \\
LINKX & 72.32{\tiny$\pm$1.41} & 78.75{\tiny$\pm$1.34} & 68.42{\tiny$\pm$1.32} & 61.81{\tiny$\pm$1.80} & 37.55{\tiny$\pm$0.36} \\
FSGNN & 67.51{\tiny$\pm$1.65} & 66.35{\tiny$\pm$1.16} & 78.27{\tiny$\pm$1.28} & 74.10{\tiny$\pm$1.89} & 79.92{\tiny$\pm$0.56} \\
$G^2$ & 76.63{\tiny$\pm$1.63} & 80.36{\tiny$\pm$1.18} & 71.40{\tiny$\pm$2.38} & 64.26{\tiny$\pm$2.38} & 82.16{\tiny$\pm$0.78}  \\
\midrule
Di-GCN & 87.36{\tiny$\pm$1.06} & 92.75{\tiny$\pm$0.75} & 52.24{\tiny$\pm$3.65} & 37.74{\tiny$\pm$1.54} & 52.71{\tiny$\pm$0.32} \\
MagNet & 85.26{\tiny$\pm$1.05} & 93.38{\tiny$\pm$0.89} & 58.32{\tiny$\pm$2.87} & 39.01{\tiny$\pm$1.93} & 88.07{\tiny$\pm$0.27} \\
Dir-GNN & 84.45{\tiny$\pm$1.69} & 92.79{\tiny$\pm$0.59} & 79.71{\tiny$\pm$1.26} & 75.31{\tiny$\pm$1.92} & \underline{91.23{\tiny$\pm$0.32}} \\
NDDGNN & \underline{88.14{\tiny$\pm$1.28}} & \underline{94.17{\tiny$\pm$0.58}} & \underline{79.79{\tiny$\pm$1.04}} & \underline{75.38{\tiny$\pm$1.95}} & \textbf{91.76{\tiny$\pm$0.27}} \\
\midrule
\textbf{DHGNN} & \textbf{88.39}{\tiny$\pm$0.98} & \textbf{94.60}{\tiny$\pm$0.52} & \textbf{80.11}{\tiny$\pm$1.73} & \textbf{76.84}{\tiny$\pm$1.61} & 89.05{\tiny$\pm$0.53} \\
\bottomrule
\end{tabular}
\label{tab:main}
\end{table}

\begin{table}[htbp]
\centering
\caption{Link prediction accuracy (\%) on homophilic and heterophilic graphs. The best results are highlighted in \textbf{bold}, while the second-best results are marked with \underline{underline}.}
\renewcommand{\arraystretch}{1.2}
\setlength{\tabcolsep}{8pt}
\scalebox{0.8}{
\begin{tabular}{l|cccc}
\toprule
\textbf{Datasets} & \textbf{Cora-ML} & \textbf{Citeseer-Full} & \textbf{Chameleon} & \textbf{Squirrel} \\
\midrule
GCN & 76.24$\pm$4.25 & 71.16$\pm$4.41 & 86.03$\pm$1.53 & 90.64$\pm$0.49 \\
GAT & 64.58$\pm$12.28 & 67.28$\pm$2.90 & 85.12$\pm$1.57 & 90.37$\pm$0.45 \\
\midrule
MagNet & 82.20$\pm$3.36 & 85.89$\pm$4.61 & 86.28$\pm$1.23 & 90.13$\pm$0.53 \\
Dir-GNN & 82.23$\pm$4.01 & 70.62$\pm$6.56 & 82.76$\pm$2.72 & 89.17$\pm$0.52 \\
NDDGNN & \underline{83.04$\pm$9.56} & \underline{87.93$\pm$3.92} & \underline{90.30$\pm$1.82} & \underline{92.40$\pm$1.01} \\
\midrule
\textbf{DHGNN} & \textbf{98.11}$\pm$1.84 & \textbf{89.70}$\pm$6.76 & \textbf{99.09}$\pm$2.02 & \textbf{96.21}$\pm$3.42 \\
\bottomrule
\end{tabular}}
\label{tab:link}
\end{table}

We evaluate our model on five datasets using the same data splits as in \citep{nddgnn2024}. We evaluate the performance by node classification accuracy with standard deviation in the semi-supervised setting. For Squirrel and Chameleon, we use 10 publicly-available splits (48\%/32\%/20\% for training/validation/testing) provided by~\citep{pei2020geom}. For the remaining datasets, we adopt the same splits as \citep{dirgnn2024}. The results, presented in Table~\ref{tab:main}, report the mean classification accuracy on the test nodes over 10 random splits. We also provide a sensitivity analysis of two hyperparameters for the node classification tasks. The corresponding results are presented in \ref{app:hyper}.

Compared to methods designed for undirected heterophilic graphs, our proposed model consistently outperforms all undirected baselines. Specifically, it achieves average absolute improvements of 11.76\%, 14.24\%, 1.84\%, 2.74\% and 6.89\% across the five datasets when compared to the strongest undirected method on each. It can be observed that MLP outperforms some of the GNN-based methods on directed datasets such as Cora-ML and Citeseer-Full. This suggests that these methods may struggle to effectively leverage structural information for node classification on directed graphs.

In addition, our model surpasses existing state-of-the-art methods tailored for directed graphs on four out of the five datasets. Take the heterophilic Squirrel dataset as an example. It exhibits the highest graph density and the lowest homophily ratio among the five datasets, indicating more complex inter-class linking patterns. The improvements on this dataset highlight our model’s strong ability to leverage inter-class connections effectively. The performance variation on Roman-Empire may be attributed to the distinct nature of this graph, which is based on semantic relationships between words rather than hyperlinks or citations. In the Roman-Empire dataset, two words are connected if they appear consecutively or are linked via the dependency tree of a sentence. This construction may reduce both the directional asymmetry and the long-range dependencies between words, thereby limiting DHGNN's ability to extract useful information for node classification. 

\subsection{Link prediction}

In addition to node classification, we assess the performance of the proposed DHGNN on link prediction tasks. In these tasks, the edges in directed graphs are treated as ordered pairs. Negative samples are generated using the negative\_sampling method in PyG~\citep{pygcite}, following the same strategy as in \citep{nddgnn2024}. These negative samples exclude both the original edges and their reversed counterparts. The evaluation metric for link prediction is accuracy. This metric is not significantly affected by class imbalance since the number of negative samples matches that of positive edges. Although the traditional definition of heterophily based on node labels is not directly applicable in link prediction tasks, DHGNN can still capture the feature discrepancy between neighboring nodes. Results are reported in Table~\ref{tab:link}. Our model achieves the highest prediction accuracy across all datasets, with absolute improvements of 15.07\%, 1.77\%, 8.79\%, and 3.81\% over the best-performing baseline on each dataset, respectively. These results indicate that the proposed method effectively captures structural information from directed neighborhoods. This may be attributed to the fact that DHGNN maintains independent encoders for each direction, rather than performing fusion after every message passing layer in NDDGNN. This design enables each encoder to capture long-range dependencies along extended paths more effectively, thereby providing richer structural information for link prediction.

\subsection{Ablation study}

\begin{table*}[ht]
\centering
\caption{Ablation study for node classification. The best result (accuracy) is highlighted in \textbf{bold}. \textbf{+ResG.} denotes the resettable homophily-aware gating mechanism for message passing, and \textbf{+Fus.} denotes the structure-aware fusion module.}
\scalebox{0.9}{
\begin{tabular}{cccc|ccc}
\toprule
\textbf{+ResG.} & \textbf{+Fus.} & \textbf{+$L_{branch}$} & \textbf{+$L_{imp}$} & \textbf{Cha.} & \textbf{Squ.} & \textbf{Rom.}\\
\midrule
- & - & - & - & 
79.21$\pm$1.03 & 74.96$\pm$1.95 & 43.35$\pm$0.63\\
\checkmark & - & - & - &
79.52$\pm$1.38 & 75.22$\pm$2.13 & 85.55$\pm$0.46\\
\checkmark & \checkmark & - & - &
79.98$\pm$1.68 & 76.38$\pm$1.57 & 86.50$\pm$0.36\\
\checkmark & \checkmark & \checkmark & - &
79.54$\pm$1.17 & 76.25$\pm$1.84 & 88.06$\pm$0.44\\
\checkmark & \checkmark & \checkmark & \checkmark &
\textbf{80.11}$\pm$1.73 & \textbf{76.84}$\pm$1.61 & \textbf{89.05}$\pm$0.53\\
\bottomrule
\end{tabular}}
\label{tab:ablation}
\end{table*}

To evaluate the effectiveness of different components in our proposed framework, we conduct an ablation study focusing on four key elements: the resettable homophily-aware gating mechanism (ResG.), the structure-aware noise-tolerant fusion module (Fus.), and the two auxiliary losses, i.e., $L_{branch}$ and $L_{imp}$. When all components are disabled (the first row in Table~\ref{tab:ablation}), the model degenerates to a two-branch vanilla GCN that encodes each direction individually, and a linear layer followed by summation as the alternative fusion module. The results in Table~\ref{tab:ablation} demonstrate that the homophily-aware gating mechanism significantly improves performance on heterophilic datasets. Additionally, the structure-aware fusion module enhances the extraction of valuable information from the directional encoders. The results also suggest that the combined use of the two auxiliary losses can lead to synergistic effects, further boosting classification accuracy.



\subsection{Visualization of gating vectors}

\begin{figure*}[htbp]
    \centering
    \begin{subfigure}[t]{0.49\textwidth}
        \includegraphics[width=\linewidth]{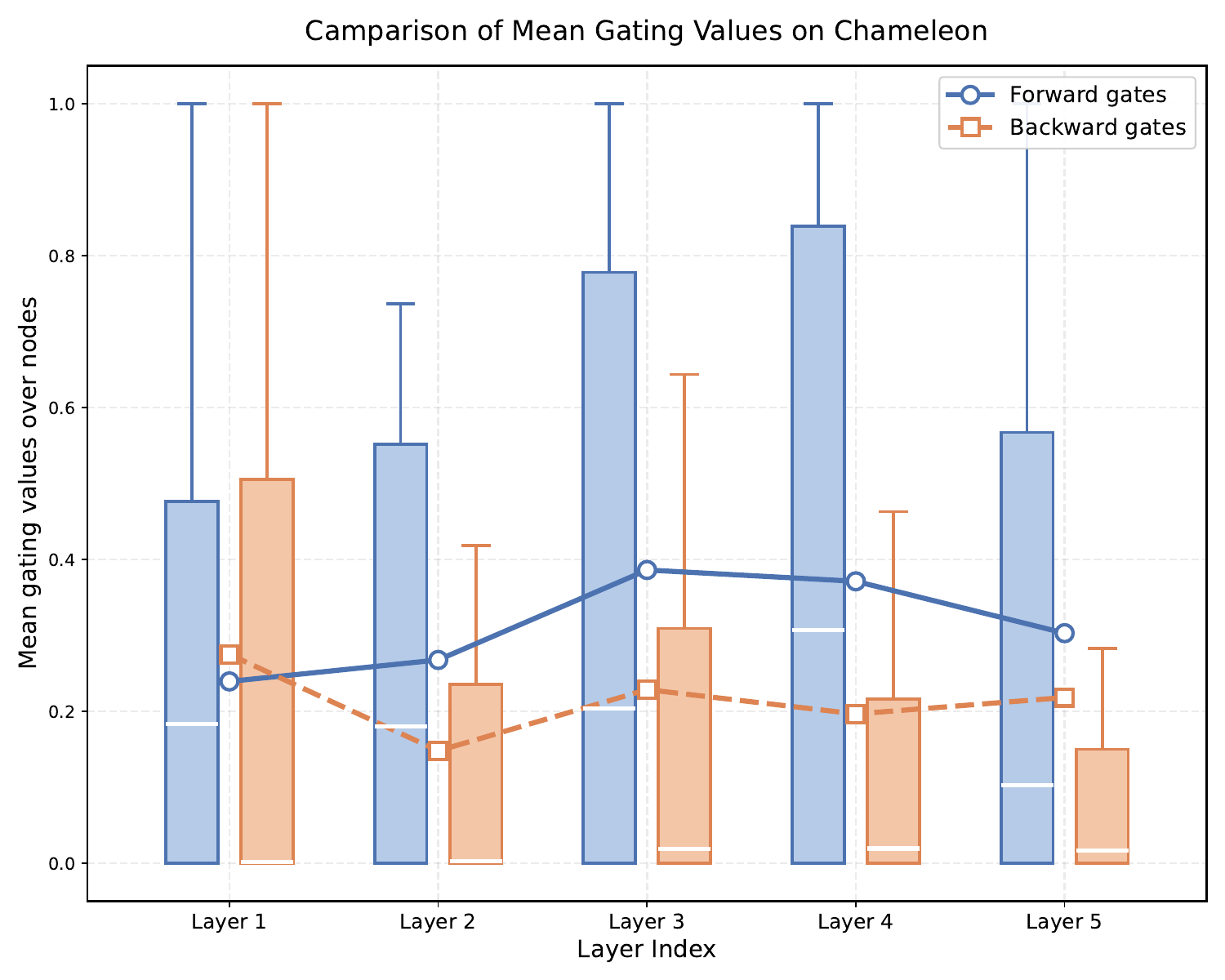}
        \caption{Forward/backward gating values on Chameleon.}
    \end{subfigure}
    \hfill
    \begin{subfigure}[t]{0.49\textwidth}
        \includegraphics[width=\linewidth]{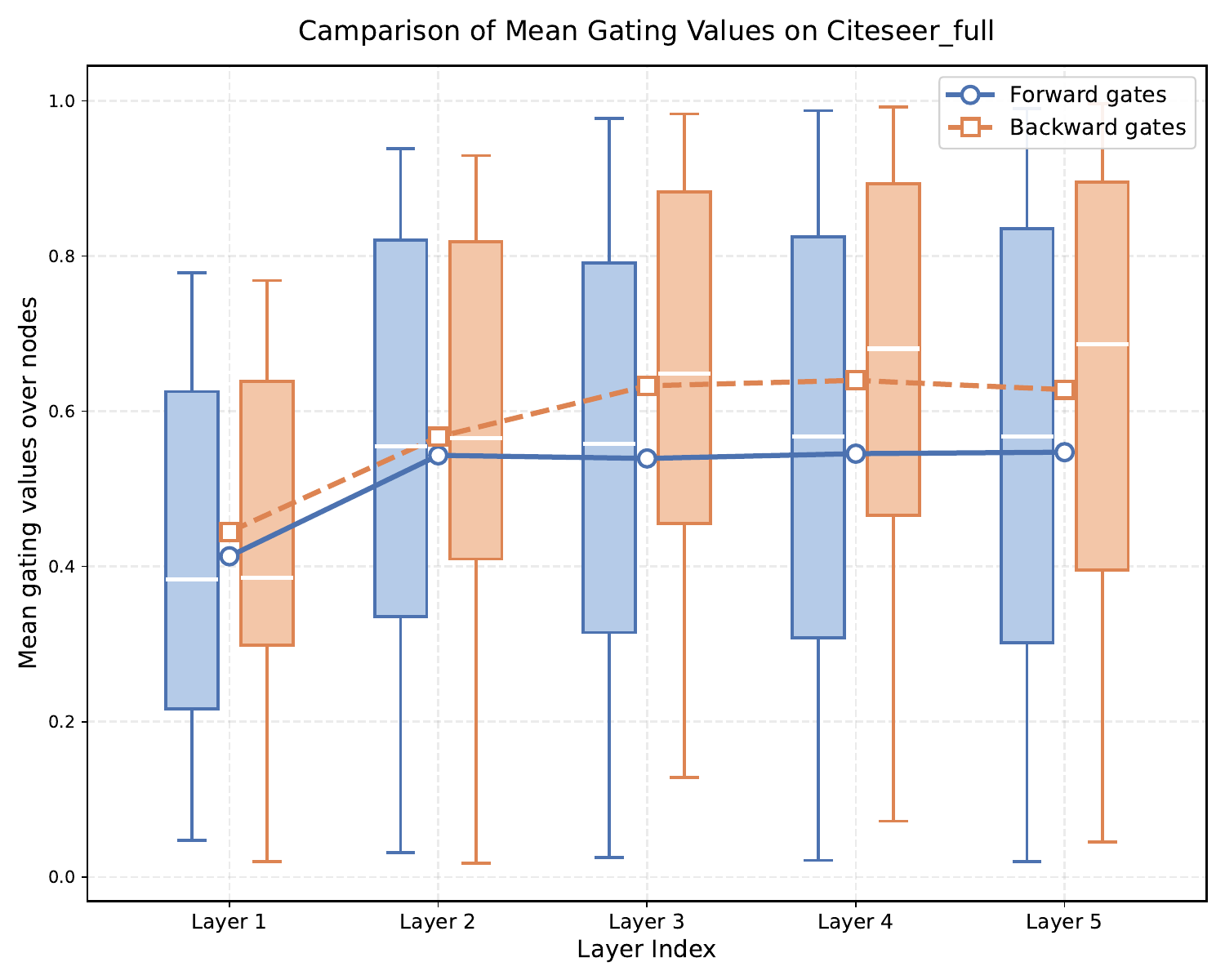}
        \caption{Forward/backward gating values on Citeseer-Full.}
    \end{subfigure}
    \caption{The gating values of the first five layers on heterophilic and homophilic datasets. The dots on line charts are the mean value over both nodes and chunks, while the box-plots present the distribution of mean gating values over chunks.}
    \label{fig:gates}
\end{figure*}

To better understand the behavior of our homophily-aware gated message passing mechanism, we visualize the gating values on two representative datasets: the heterophilic Chameleon and the homophilic Citeseer-Full, as shown in Figure~\ref{fig:gates}. A clear gap in gating values between the two directions is observed on Chameleon, highlighting a strong directional asymmetry. While a similar trend is present on Citeseer-Full, the gap is notably smaller, suggesting that the gating mechanism adaptively responds to the directional differences in homophily on different types of graphs.

Additionally, we observe a decrease in gating values of the backward direction from layer 1 to layer 2 on the Chameleon dataset. This aligns with the higher homophily ratio at the second hop, as illustrated in Figure~\ref{subfig:chameleon}. The larger variance in gating values, as indicated by the taller boxplots, may be attributed to the increased variability in neighborhood homophily at greater hop distances. These findings indicate that the proposed gating mechanism not only distinguishes between directional neighborhoods but also dynamically adapts to the non-monotonic variations in homophily across layers. Similar gating behavior is observed on other datasets, as visualized in \ref{app:gate}.

\subsection{Alleviating Over-smoothing in DHGNN}
\begin{figure*}[htbp]
    \centering
    \begin{subfigure}[t]{0.45\textwidth}
        \includegraphics[width=\linewidth]{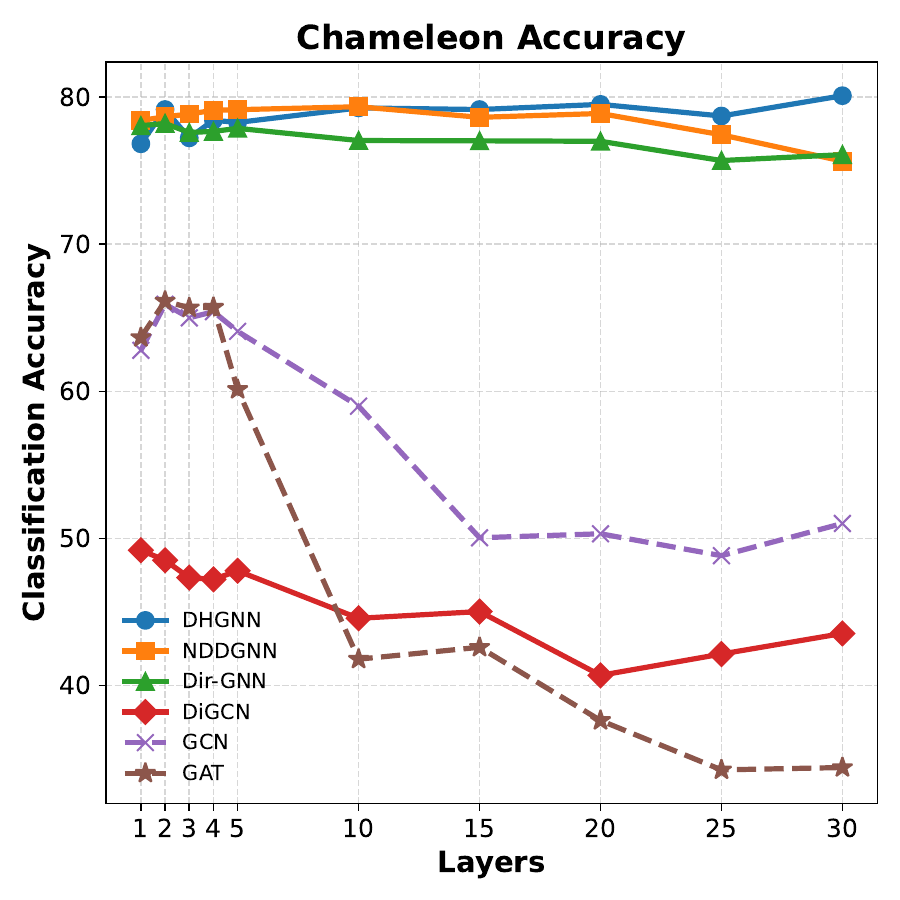}
        \caption{Node classification accuracy on Chameleon at deeper layers, better visualized in color.}
        \label{fig:chameleonsmooth}
    \end{subfigure}
    \hfill
    \begin{subfigure}[t]{0.49\textwidth}
        \includegraphics[width=\linewidth]{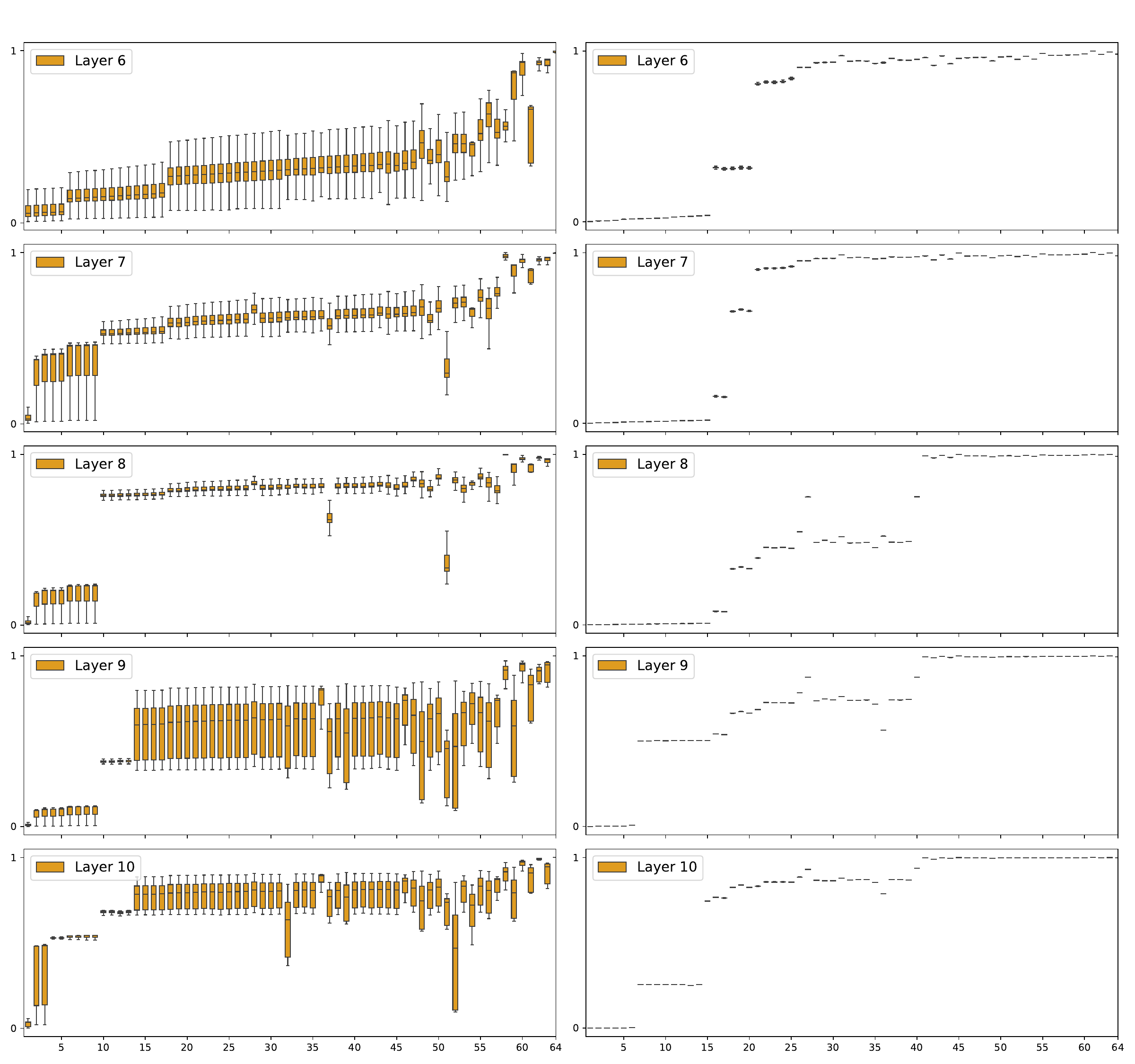}
        \caption{Gating values on Chameleon at deeper layers.}
        \label{fig:chameleondeep}
    \end{subfigure}
    \caption{The node classification accuracy and behavior of gating values at deeper layers.}
    \label{fig:deep}
\end{figure*}

As shown in Figure~\ref{fig:chameleonsmooth}, DHGNN effectively mitigates the over-smoothing problem. Node classification accuracy does not decrease, and even slightly improves, as the model goes deeper. The change in accuracy is obviously smaller than that observed in traditional GNNs, and remains lower than other baselines. Analysis of the gating vectors at deeper layers (Figure~\ref{fig:chameleondeep}) shows that while some gates saturate and refuse neighborhood messages, others remain open with lower values. This suggests that the model continues to capture informative signals from distant nodes, even at greater depths. We visualize the results on additional datasets in \ref{app:oversmoothing}, which exhibit similar patterns to those in Figure~\ref{fig:chameleondeep}, further suggesting that DHGNN effectively alleviates over-smoothing on heterophilic graphs.

\subsection{Runtime Evaluation}

To examine the efficiency of DHGNN, we compare it with the GCN-like baseline NDDGNN \citep{nddgnn2024}. Both models use 4 layers and the same hidden dimension. The reported time in Table~\ref{tab:runtime} is the average per-epoch cost, including training and validation, measured in seconds.

\begin{table}[htbp]
\centering
\caption{Comparison of runtime per-epoch (in seconds).}
\begin{tabular}{c|ccccc}
\toprule
& \textbf{Cora.} & \textbf{Citeseer.} & \textbf{Chameleon} & \textbf{Squirrel} & \textbf{Roman.}\\
\midrule
NDDGNN & 0.0634 & 0.0529 & 0.0643 & 0.1267 & 0.1430 \\
DHGNN & 0.0455 & 0.1191 & 0.1026 & 0.1165 & 0.2090 \\
\bottomrule
\end{tabular}
\label{tab:runtime}
\end{table}

The training time of DHGNN is on the same order of magnitude as NDDGNN. Although it is slightly higher on some datasets, it remains efficient and even faster on the largest graph, Squirrel, indicating that the gating mechanism introduces only moderate computational overhead.

\section{Conclusion}

We introduced Directed Homophily-aware Graph Neural Network (DHGNN), a novel framework that adaptively integrates neighborhood information by accounting for both homophily gaps between opposite directions and homophily fluctuations across layers. DHGNN features a homophily-aware gating mechanism with a resettable update strategy and a structure-aware, noise-tolerant fusion module that selectively emphasizes informative directional representations. Comprehensive experiments on both homophilic and heterophilic datasets show that DHGNN can outperform state-of-the-art methods on node classification and link prediction tasks. Gating analyses further confirm its ability to adaptively respond to complex homophily patterns. Our method not only advances effective learning on directed graphs but also introduces a robust mechanism for extracting informative signals from obscure or noisy neighborhoods.

While DHGNN demonstrates to be effective, its scalability on extremely large graphs and generalization to broader domains beyond citation and web graphs remain promising directions for future exploration. We aim to further enhance the model’s efficiency and extend its applicability across diverse graph domains, paving the way for broader adoption in real-world scenarios.

\section{Acknowledgment}

This research/project is supported by the Ministry of Education, Singapore under its Academic Research Fund Tier 1 (RG16/24). Any opinions, findings and conclusions or recommendations expressed in this material are those of the author(s) and do not reflect the views of the Ministry of Education, Singapore.

\appendix
\setcounter{table}{0}
\setcounter{figure}{0}
\renewcommand{\thetable}{A\arabic{table}}
\renewcommand{\thefigure}{A\arabic{figure}}

\section{Analysis of Homophily Ratio with SBM}
\label{app:homoanalysis}

To illustrate that homophily may increase with hop distance, we consider a directed two-block SBM with classes $A, B$ (balanced with \( n \) nodes in each class for simplicity). Each directed edge $(i,j)$ appears independently with:
\[
\mathbb{P}[i \!\to\! j] =
\begin{cases}
p_{\text{in}}, & \text{if } \mathrm{class}(i)=\mathrm{class}(j),\\[3pt]
p_{\text{out}}, & \text{otherwise.}
\end{cases}
\]
The expected one-hop homophily is 
\[
h_1(u) = \frac{(n-1)p_{\text{in}}}{(n-1)p_{\text{in}} + n p_{\text{out}}}.
\]
When \( p_{\text{out}} > p_{\text{in}} \), this value is small, indicating a heterophilous 1-hop neighborhood.

To compute the 2-hop homophily, we consider all 2-hop paths \( u \to w \to v \), where \( w \) is a 1-hop neighbor and \( v \) is a 2-hop neighbor of \( u \). Let \( h_2(u) \) denote the probability that a randomly selected 2-hop neighbor \( v \) shares the same class as \( u \). This can be decomposed based on the class of the intermediate node \( w \). The probability that a 1-hop neighbor \( w \) belongs to class \( A \) is:
\[
\mathbb{P}(w \in A \mid u \in A) = \frac{(n - 1) p_{\text{in}}}{(n - 1) p_{\text{in}} + n p_{\text{out}}},
\]
while \( w \in B \) with probability:
\[
\mathbb{P}(w \in B \mid u \in A) = \frac{n p_{\text{out}}}{(n - 1) p_{\text{in}} + n p_{\text{out}}}.
\]
Conditional on \( w \)'s class, the probability that the 2-hop neighbor \( v \in A \) is given by:
\begin{align*}
\mathbb{P}(v \in A \mid w \in A) = \frac{(n - 1) p_{\text{in}}}{(n - 1) p_{\text{in}} + n p_{\text{out}}}, \\
\mathbb{P}(v \in A \mid w \in B) = \frac{n p_{\text{out}}}{(n - 1) p_{\text{in}} + n p_{\text{out}}}.
\end{align*}
Thus, the 2-hop homophily is:
\begin{align*}
h_2(u) = \;
& \left(\frac{(n - 1) p_{\text{in}}}{(n - 1) p_{\text{in}} + n p_{\text{out}}}\right)^2
\\
+ \;
&  \left(\frac{n p_{\text{out}}}{(n - 1) p_{\text{in}} + n p_{\text{out}}}\right)^2.
\end{align*}
When \( p_{\text{out}} > p_{\text{in}} \), for instance, if we fix the class size $n=500$, the intra-class and inter-class connection probabilities $p_{\text{in}}=0.2$ and $p_{\text{out}}=0.8$, the two homophily ratios $h_1(u)\text{, and } h_2(u)$ are 0.1999 and 0.6804 respectively. Under this setting, \( h_2(u) \) exceeds \( h_1(u) \), indicating that homophily may increase at the 2-hop level. This illustrates that homophily is not necessarily a monotonically decreasing function of hop distance and suggests that the design of GNNs for heterophilic graphs should consider the fluctuations of homophily ratio in the aggregation of multiple layers.

\section{Monotonicity Analysis of the Homophily-Aware Resettable Gate}
\label{app:monoanalysis}

Consider the same directed SBM model, but now with class sizes $n_A$ and $n_B$. To keep algebra transparent, we adopt scalar embeddings $\mathbf{h}_v^{(l)}\in \mathbb{R}$ for each node. Initially $\mathbf{h}_v^{(0)}=+1$ if $v\in A$, and $\mathbf{h}_v^{(0)}=-1$ if $v\in B$. Under such settings, the expected one-hop homophily is 
\[
h_1 = \frac{(n_A-1)p_{\text{in}}}{(n_A-1)p_{\text{in}} + n_B p_{\text{out}}}.
\]

\begin{lemma}[Expected neighbor average]
For $v\!\in\!A$ under mean-field approximation, the $l$-hop message can be expressed as:
\[
\mathbb{E}[\mathbf{m}_v^{(l)}] = (+1)\cdot h_l +(-1) \cdot (1- h_l)= 2h_l - 1, 
\]

\end{lemma}

\noindent
where $h_l$ is the $l$-hop homophily defined by the proportion of nodes in class A in the $l$-hop neighbors of the central node.

Suppose the similarity between the center node $v$ and its $l$-hop neighborhood is defined by innner product between the node representation and the current message:

\[
\mathbb{E}\!\left[s_v^{(l)}\right] = \mathbb{E} \left[ \mathbf{x}_v \, \mathbf{m}_v^{(l)} \right]=  2h_l - 1.
\]

The same result can be derived for nodes in class $B$.

\noindent
Assume that the gate proposal at layer $l$ is an affine decreasing function of the node-message similarity, then it can also be expressed as an affine decreasing function of the expected homophily ratio $h_l$:

\[
\hat{\mathbf{g}}_v^{(l)} \;=\; \alpha_1 -\alpha_2\, s_v^{(l)}=\alpha_1 + \alpha_2 - 2\,\alpha_2 h_l  \qquad \alpha_2>0,
\]

\noindent
where $\alpha_1, \alpha_2$ are weights of the affine function.

\begin{theorem}[Gate Non-Monotonicity]
In DHGNN, there exist graph structures and parameter configurations where the gate sequence $\{\tilde{\mathbf{g}}_v^{(l)}\}_{l=0}^L$ is non-monotonic.
\end{theorem}

\begin{proof}
Consider a directed SBM with parameters: $n=500$, $p_{\text{in}}=0.2$, $p_{\text{out}}=0.8$, yielding homophily ratios $h_1 \approx 0.20$, $h_2 \approx 0.68$. Without loss of generality, we assume $\alpha_1=\alpha_2=1/2$ and $\hat{\mathbf{g}}_v^{(l)} = \mathbf{r}_v^{(l)} = 1 - h_l$ in the following analysis:

\noindent
\textbf{Layer 1:}
\begin{itemize}
    \item High heterophily ($h_1 \approx 0.20$) $\Rightarrow$ Small $s_v^{(l)}$, $\hat{\mathbf{g}}_v^{(1)} \approx 0.80$
    \item With $\tilde{\mathbf{g}}_v^{(0)} = 0$, we have $\hat{\mathbf{g}}_v^{(1)} > \tilde{\mathbf{g}}_v^{(0)}$
    \item Reset gate $\mathbf{r}_v^{(1)} \approx 0.80$ $\Rightarrow$ $\tilde{\mathbf{g}}_v^{(1)} \approx 0.80 \times 0.80 + 0.20 \times 0 = 0.64$
\end{itemize}

\noindent
\textbf{Layer 2:}
\begin{itemize}
    \item Higher homophily ($h_2 \approx 0.68$) $\Rightarrow$ Large $s_v^{(l)}$, $\hat{\mathbf{g}}_v^{(2)} \approx 0.32$
    \item With $\tilde{\mathbf{g}}_v^{(1)} \approx 0.64$, we have $\hat{\mathbf{g}}_v^{(2)} < \tilde{\mathbf{g}}_v^{(1)}$
    \item Reset gate $\mathbf{r}_v^{(2)} \approx 0.32$ $\Rightarrow$ $\tilde{\mathbf{g}}_v^{(2)} \approx 0.32 \times 0.32 + 0.68 \times 0.64 = 0.5376$
\end{itemize}

\noindent 
This demonstrates:
\begin{itemize}
    \item $\tilde{\mathbf{g}}_v^{(1)} > \tilde{\mathbf{g}}_v^{(0)}$ (increase from layer 0 to 1)
    \item $\tilde{\mathbf{g}}_v^{(2)} < \tilde{\mathbf{g}}_v^{(1)}$ (decrease from layer 1 to 2)
\end{itemize}
Hence, the gate sequence is non-monotonic.
\end{proof}

\noindent
\textbf{Reset gate dynamics.} The reset gate $\mathbf{r}_v^{(l)}$ controls the interpolation between preservation and adaptation:

\[
\tilde{\mathbf{g}}_v^{(l)} = (1 - \mathbf{r}_v^{(l)}) \circ \tilde{\mathbf{g}}_v^{(l-1)} + \mathbf{r}_v^{(l)} \circ \hat{\mathbf{g}}_v^{(l)}
\]

The gate difference between consecutive layers is:
\[
\Delta\tilde{\mathbf{g}}_v^{(l)} = \tilde{\mathbf{g}}_v^{(l)} - \tilde{\mathbf{g}}_v^{(l-1)} = \mathbf{r}_v^{(l)} \circ \left(\hat{\mathbf{g}}_v^{(l)} - \tilde{\mathbf{g}}_v^{(l-1)}\right)
\]

Since $\mathbf{r}_v^{(l)} \geq 0$ element-wise, the sign of $\Delta\tilde{\mathbf{g}}_v^{(l)}$ is determined by $\text{sign}(\hat{\mathbf{g}}_v^{(l)} - \tilde{\mathbf{g}}_v^{(l-1)})$.

\begin{lemma}[Reset Gate Bounds]
The reset gate modulates but cannot reverse the direction of gate change:
\begin{itemize}
    \item If $\hat{\mathbf{g}}_v^{(l)} > \tilde{\mathbf{g}}_v^{(l-1)}$, then $\tilde{\mathbf{g}}_v^{(l)} \geq \tilde{\mathbf{g}}_v^{(l-1)}$
    \item If $\hat{\mathbf{g}}_v^{(l)} < \tilde{\mathbf{g}}_v^{(l-1)}$, then $\tilde{\mathbf{g}}_v^{(l)} \leq \tilde{\mathbf{g}}_v^{(l-1)}$
    \item Equality occurs only when $\mathbf{r}_v^{(l)} = 0$
\end{itemize}
\end{lemma}




This analysis demonstrates that DHGNN's resettable gating mechanism enables non-monotonic gate behavior in response to fluctuating homophily patterns across layers. The reset gate $\mathbf{r}_v^{(l)}$ provides adaptive control over this behavior, allowing the model to effectively handle complex directed graphs with varying homophily structures.

\section{Hyperparameter Settings}
\label{app:set}




The detailed settings of the hyperparameters in both the encoders and the fusion mechanism of the proposed DHGNN are presented in Table~\ref{tab:hyper}.

\begin{table*}[htbp]
\centering
\footnotesize
\caption{Hyperparameter settings for different datasets.}
\begin{tabular}{l|rrrrr}
\toprule
\textbf{Hyperparameter} & \textbf{Cor.} & \textbf{Cit.} & \textbf{Cha.} & \textbf{Squ.} & \textbf{Rom.} \\
\midrule
lr              & 0.001     & 0.005    & 0.05     & 0.01  & 0.005      \\
wd              & 1e-6      & 0.0005   & 0.0001   & 0.0001   & 5e-6     \\
GNN layers      & 4         & 14       & 12       & 5        & 6        \\
hidden dim & 256 & 256 & 256 & 256 & 256 \\
gate\_mlp layers      & 2         & 2        & 3        & 2        & 2         \\
adj\_mlp layers & 2         & 2        & 4        & 2        & 4         \\
input\_fc dropout   & 0.5       & 0.5      & 0.5      & 0.5      & 0.3       \\
dropout rate  & 0.2       & 0.0      & 0.3      & 0.3      & 0.1       \\
adj\_coef $\beta$      & 0         & 0.4      & 0.5      & 0.5      & 0         \\
imp\_coef $\lambda_1$    & 1e-3      & 1e-7     & 1e-5     & 1e-7     & 1e-3     \\
branch\_coef $\lambda_2$      & 0.9      & 0.95     & 0.95     & 0.8      & 0.95      \\
chunk size & 64 & 64 & 64 & 64 & 64 \\
\bottomrule
\end{tabular}
\label{tab:hyper}
\end{table*}

\section{Hyperparameter Sensitivity Analysis}
\label{app:hyper}

\begin{figure*}[htbp]
    \centering
    \begin{subfigure}[t]{0.49\textwidth}
        \includegraphics[width=\linewidth]{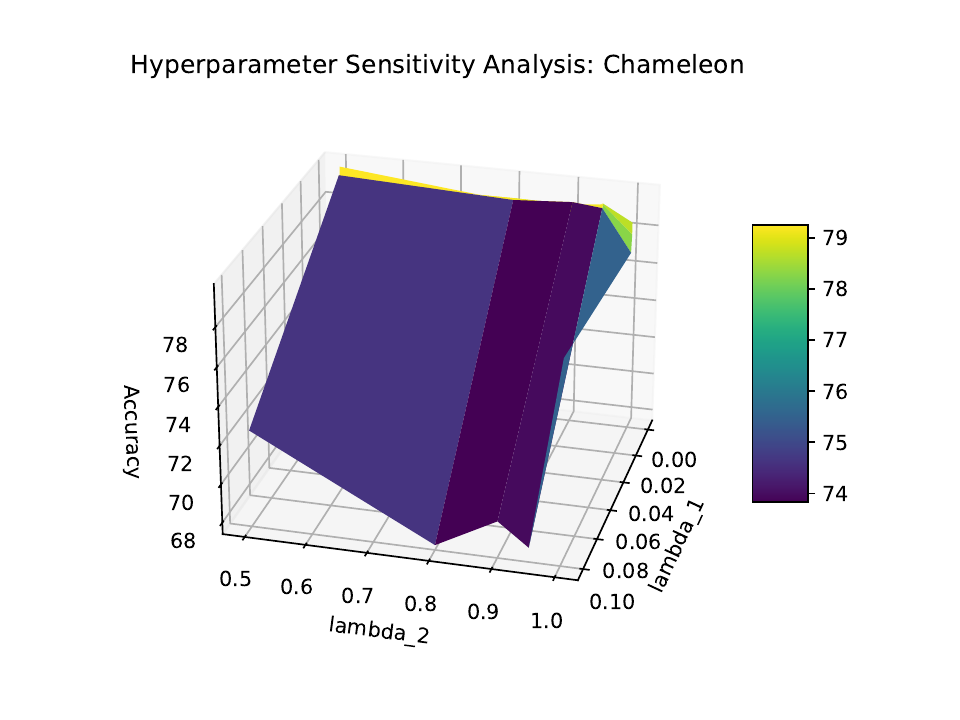}
        \caption{Hyperparameter sensitivity analysis on Chameleon.}
        \label{fig:hypercha}
    \end{subfigure}
    \hfill
    \begin{subfigure}[t]{0.49\textwidth}
        \includegraphics[width=\linewidth]{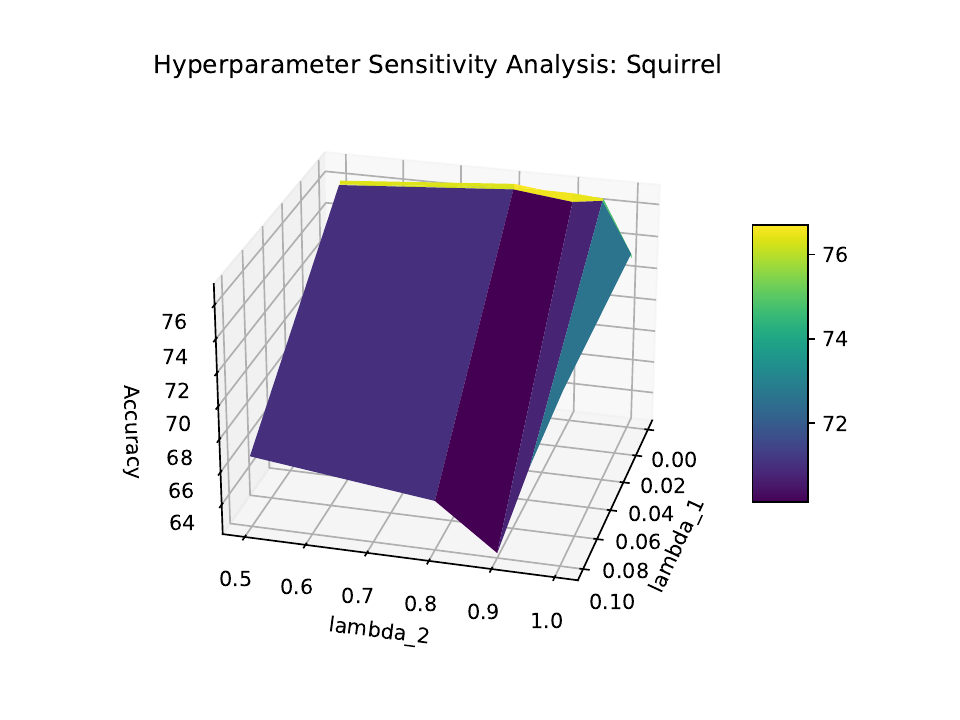}
        \caption{Hyperparameter sensitivity analysis on Squirrel.}
        \label{fig:hypersqu}
    \end{subfigure}
    \begin{subfigure}[t]{0.49\textwidth}
        \includegraphics[width=\linewidth]{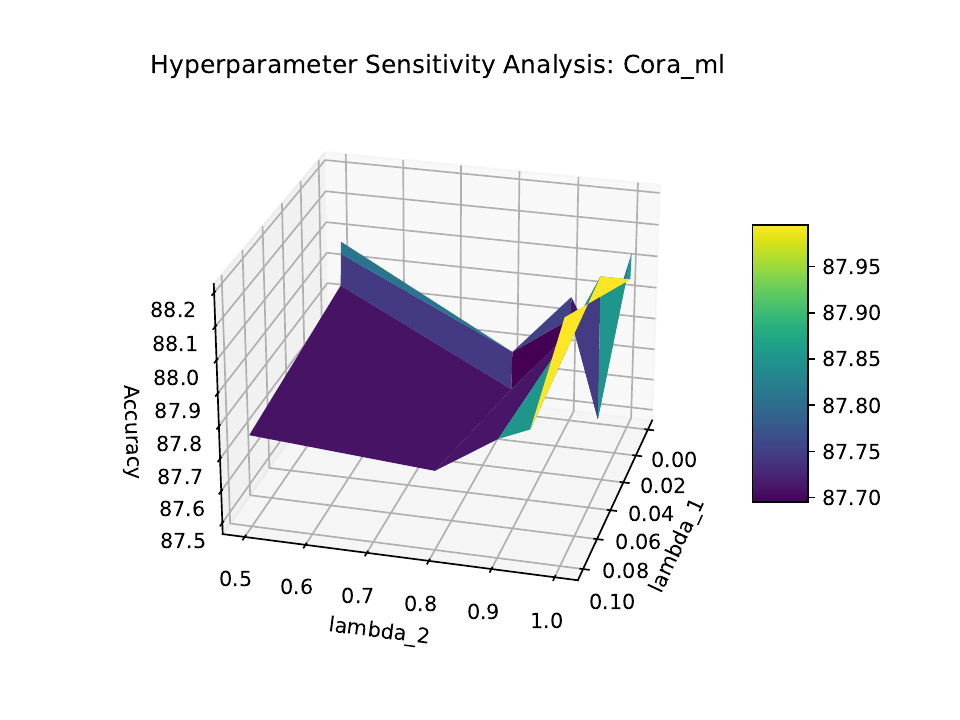}
        \caption{Hyperparameter sensitivity analysis on Cora\_ML.}
        \label{fig:hypercor}
    \end{subfigure}
    \begin{subfigure}[t]{0.49\textwidth}
        \includegraphics[width=\linewidth]{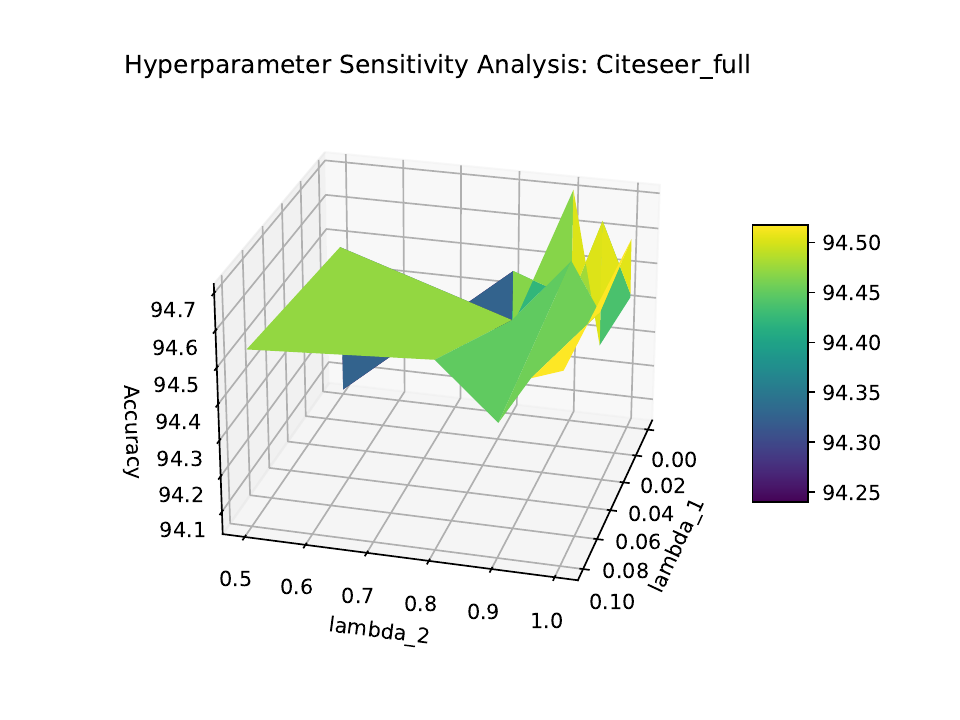}
        \caption{Hyperparameter sensitivity analysis on Citeseer\_Full.}
        \label{fig:hypercit}
    \end{subfigure}
    \caption{Hyperparameter sensitivity analysis on various datasets.}
    \label{fig:hyper}
\end{figure*}

To further understand the robustness and sensitivity of our model, we provide a hyperparameter sensitivity analysis in this section. Figure~\ref{fig:hyper} illustrates the performance variations under different settings of $\lambda_1$ and $\lambda_2$.

On the Chameleon and Squirrel datasets, the best performance is achieved when $\lambda_1$ is close to 0, indicating that the model favors embeddings from a single direction on these heterophilic graphs and tends to assign a larger weight during fusion. As a result, the constraint enforcing balanced decision scores can be relaxed in this case.

For the hyperparameter $\lambda_2$, the best node classification performance is typically achieved when its value is close to 1, suggesting that the final representation is more reliably derived from the higher-quality single-directional embeddings. This also implies that the post-fusion classification loss plays a relatively minor role in the overall optimization.

From the visualization, we may conclude that, although the best performance is sometimes achieved when $\lambda_1 = 0$, indicating the exclusion of the corresponding importance loss, both the importance loss and the branch classification loss still contribute meaningfully across different datasets. The hyperparameter analysis further confirms the effectiveness of both auxiliary losses.






\section{Visualization of the Gating Vectors on the Other Datasets}
\label{app:gate}

\begin{figure*}[htbp]
    \centering
    \begin{subfigure}[t]{0.49\textwidth}
        \includegraphics[width=\linewidth]{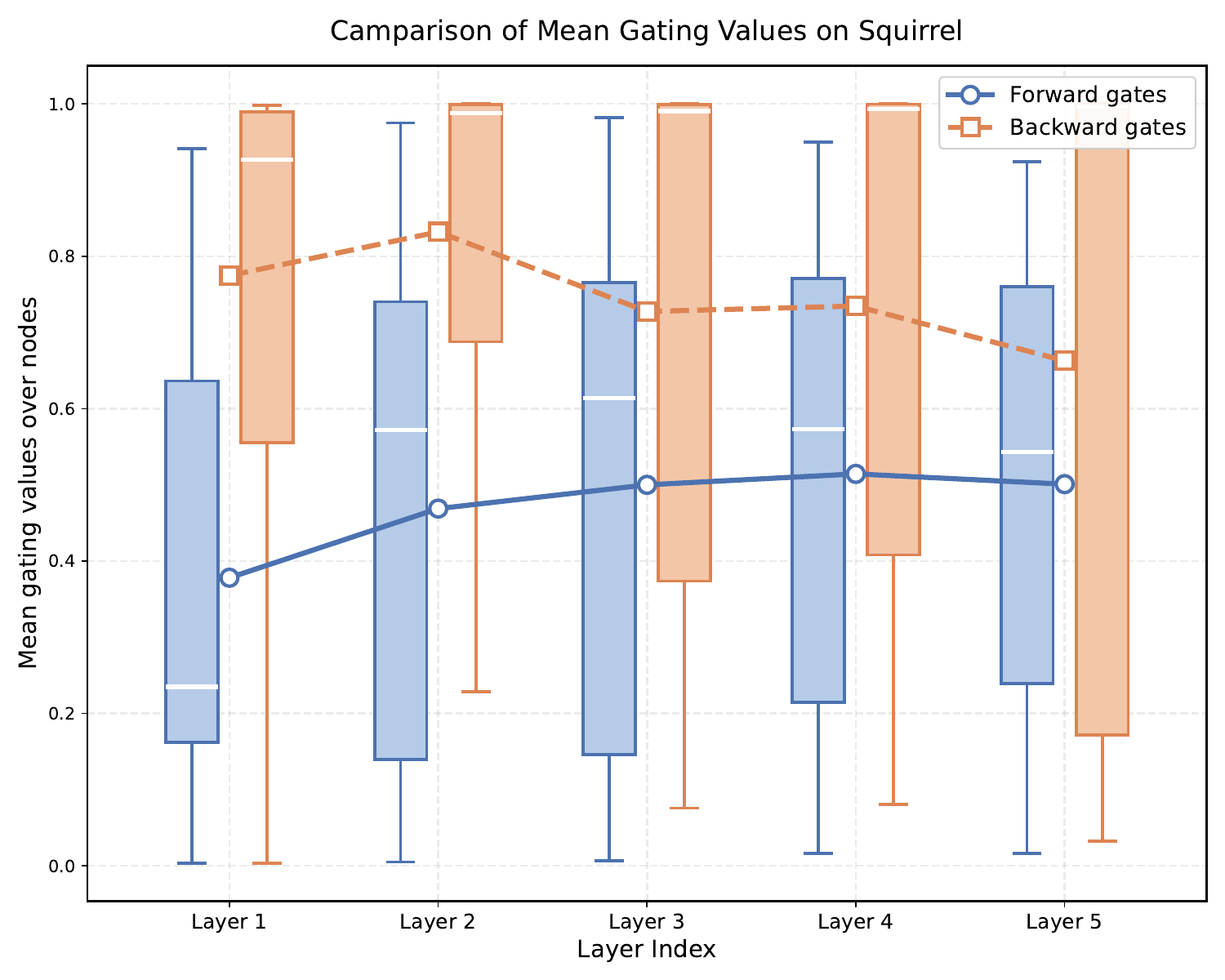}
        \caption{Forward/backward gating values on Squirrel.}
    \end{subfigure}
    \hfill
    \begin{subfigure}[t]{0.49\textwidth}
        \includegraphics[width=\linewidth]{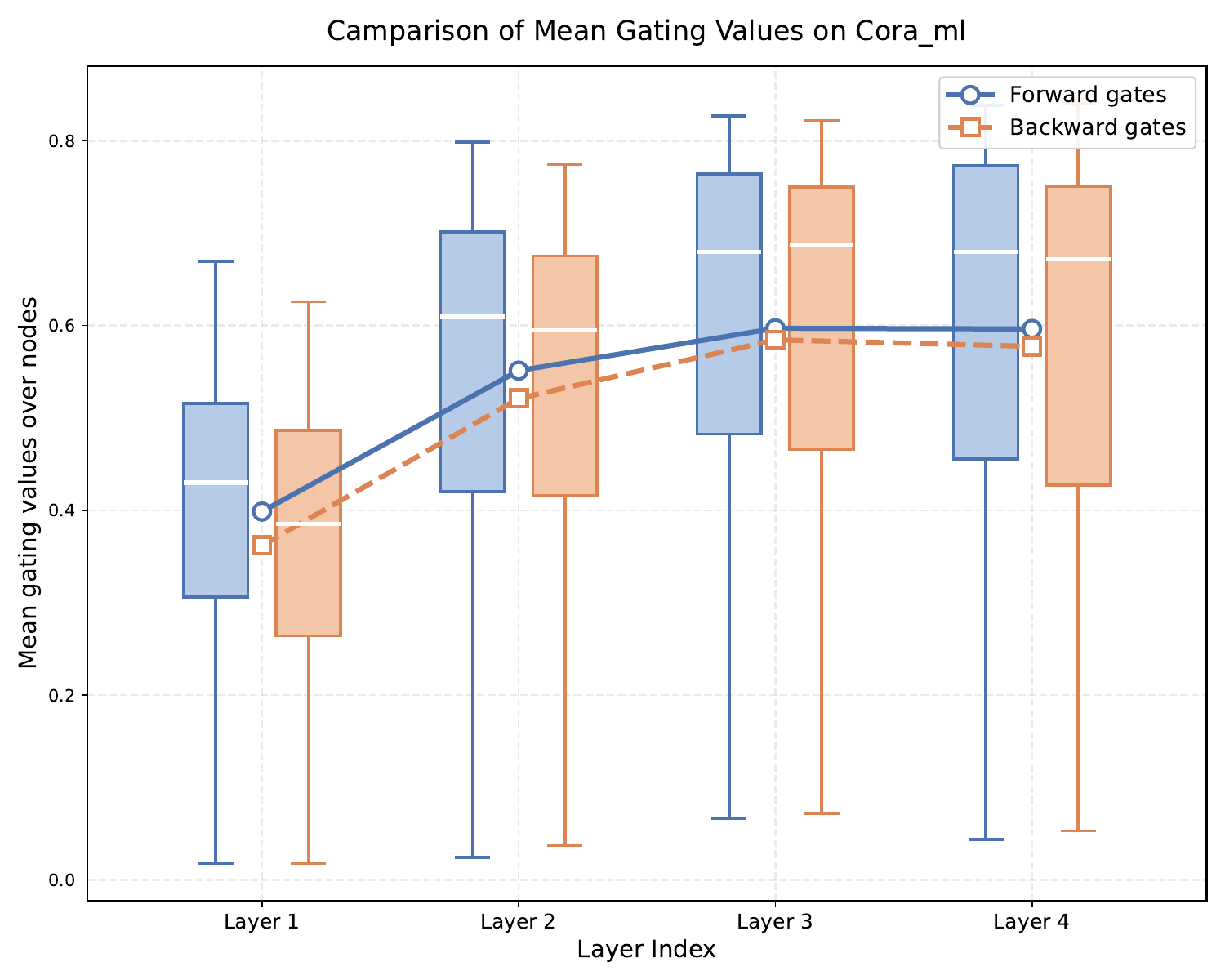}
        \caption{Forward/backward gating values on Cora-ML.}
    \end{subfigure}
    \hfill
    \begin{subfigure}[t]{0.49\textwidth}
        \includegraphics[width=\linewidth]{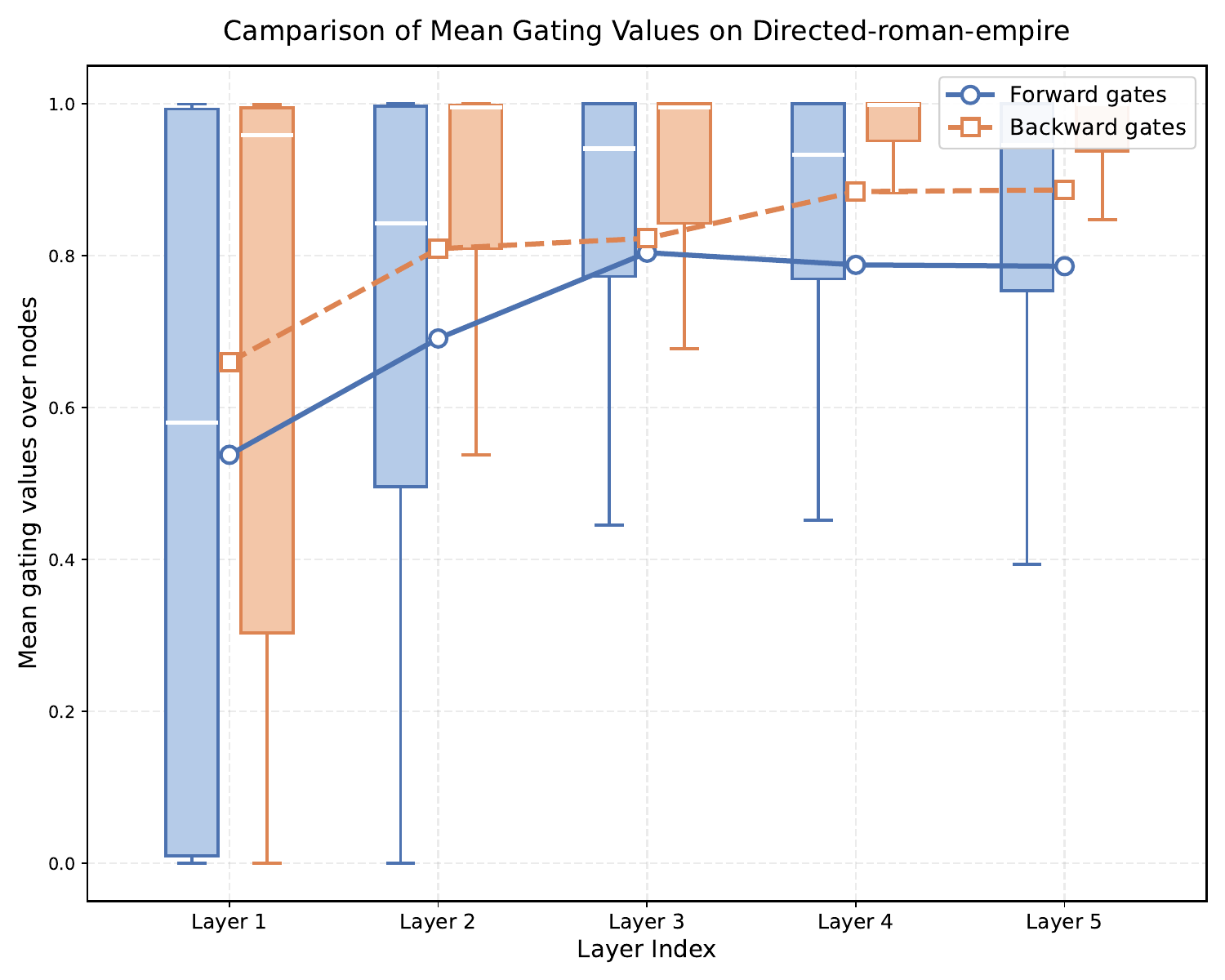}
        \caption{Forward/backward gating values on Roman-Empire.}
    \end{subfigure}
    \caption{The gating values of the first five layers on the other datasets. The dots on line charts are the mean value over both nodes and chunks, while the box-plots present the distribution of mean gating values over chunks.}
    \label{fig:gates2}
\end{figure*}

\begin{figure*}[htbp]
    \centering
    \begin{subfigure}[t]{0.45\textwidth}
        \includegraphics[width=\linewidth]{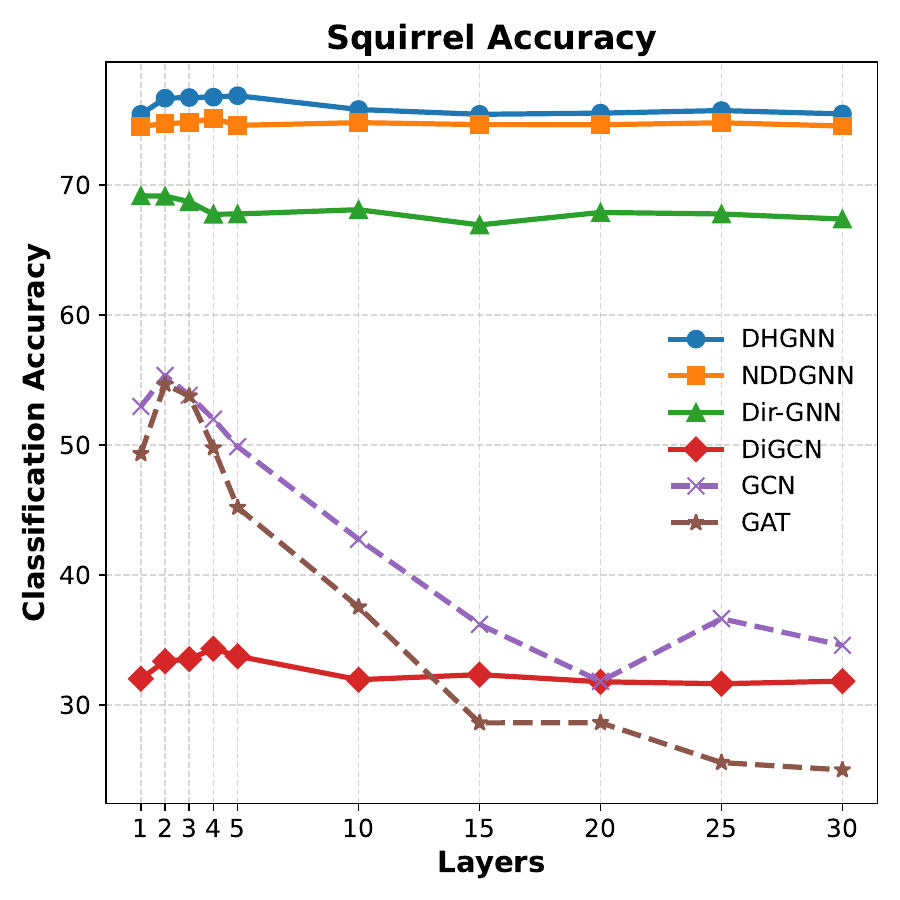}
        \caption{Node classification accuracy on Squirrel at deeper layers, better visualized in color.}
        \label{fig:squirrelsmooth}
    \end{subfigure}
    \hfill
    \begin{subfigure}[t]{0.45\textwidth}
        \includegraphics[width=\linewidth]{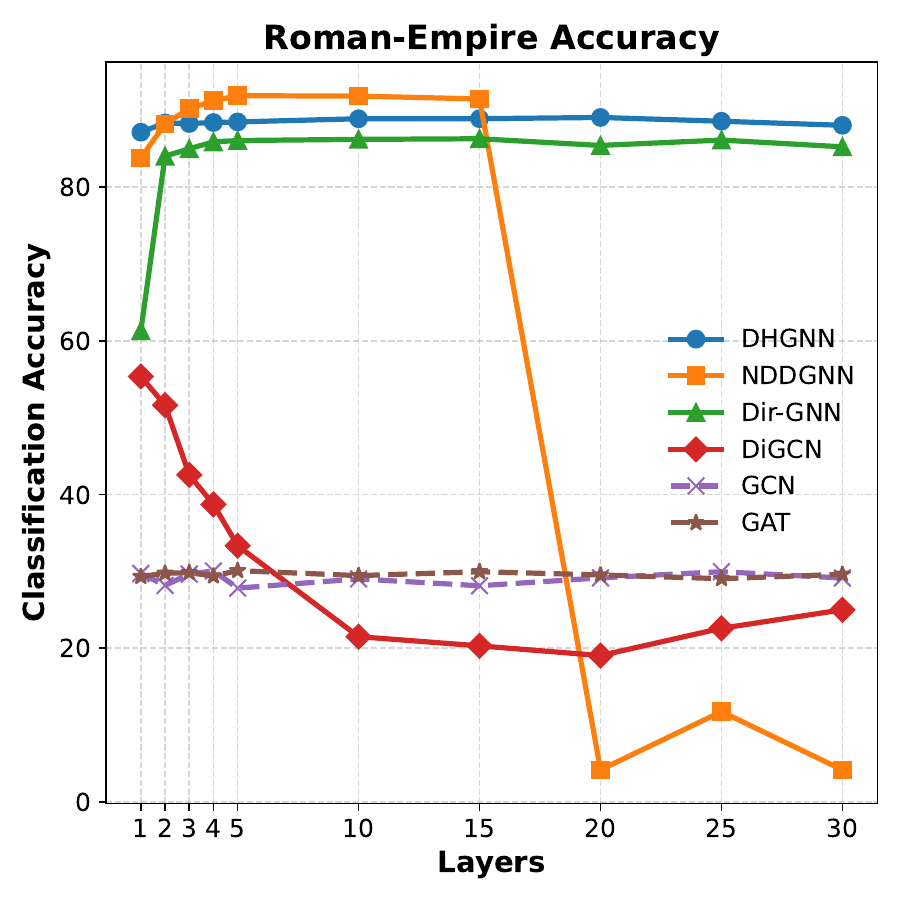}
        \caption{Node classification accuracy on Roman Empire at deeper layers, better visualized in color.}
        \label{fig:romansmooth}
    \end{subfigure}
    \caption{The node classification accuracy on other heterophilic datasets at deeper layers.}
    \label{fig:deepadd}
\end{figure*}

In addition to the visualization in Figure~\ref{fig:gates}, Figure~\ref{fig:gates2} presents gating values from other datasets, further illustrating the directional variance and the changes across different layers. 

On both the Squirrel and Roman Empire datasets, we observe some decreasing cases in the gating values, suggesting the presence of meaningful information from more distant hops. 

In contrast, for datasets where embeddings from the two directions are highly similar, such as Cora\_ML, the gating mechanism adapts accordingly, exhibiting consistent behavior across layers and learning from neighborhood information in a balanced manner.

\section{Node Classification Results on Other Datasets at Deeper Layers}
\label{app:oversmoothing}

Beyond the visualization in Figure~\ref{fig:deep}, Figure~\ref{fig:deepadd} shows the accuracy curves for node classification on other heterophilic datasets. From these two curves, we observe that the node classification performance of both DHGNN and Dir-GNN remains relatively stable as the number of layers increases, with DHGNN consistently achieving higher accuracy. NDDGNN also demonstrates stable performance on the Squirrel dataset, but its accuracy drops sharply beyond 20 layers on Roman Empire, possibly due to instability in directional fusion at each layer, leading to node embeddings that converge to indistinguishable representations. These results indicate that DHGNN effectively mitigates the over-smoothing problem on heterophilic graphs.






\clearpage
\bibliographystyle{elsarticle-num}
\bibliography{references}

\end{document}